\documentclass[lettersize,journal]{IEEEtran}
\usepackage{amsmath,amsfonts}
\usepackage{algorithmic}
\usepackage{algorithm}
\usepackage{array}
\usepackage[caption=false,font=normalsize,labelfont=sf,textfont=sf]{subfig}
\usepackage{textcomp}
\usepackage{stfloats}
\usepackage{url}
\usepackage{verbatim}
\usepackage{graphicx}
\usepackage{cite}
\usepackage{booktabs}
\usepackage{multirow}
\usepackage[table,dvipsnames]{xcolor}
\usepackage{orcidlink}
\hypersetup{hidelinks}
\definecolor{tabhighlight}{HTML}{e5e5e5}
\definecolor{codecomment}{HTML}{008000}

\begin{document}

\title{ICPL-ReID: Identity-Conditional Prompt Learning \\
for Multi-Spectral Object Re-Identification}

\author{Shihao Li$^{\orcidlink{0009-0001-0923-3965}}$, Chenglong Li$^{\orcidlink{0000-0002-7233-2739}}$, Aihua Zheng\(^{*\orcidlink{0000-0002-9820-4743}}\), Jin Tang$^{\orcidlink{0000-0001-8375-3590}}$, Bin Luo$^{\orcidlink{0000-0002-1414-3307}}$, \textit{Senior Member, IEEE}
\thanks{
This research is supported in part by the National Natural Science Foundation of China under Grants 62372003 and 62376004, the University Synergy Innovation Program of Anhui Province under Grant GXXT-2022-036, the Natural Science Foundation of Anhui Province under Grants 2308085Y40 and 2208085J18, and the Open Research Project of the Anhui Provincial Key Laboratory of Security Artificial Intelligence under Grant SAI202401. (\(^*\)The corresponding author is Aihua Zheng.)
}
\thanks{
C. Li, A. Zheng, and S. Li are with the Information Materials and Intelligent Sensing Laboratory of Anhui Province, Anhui Provincial Key Laboratory of Security Artificial Intelligence, School of Artificial Intelligence, Anhui University, Hefei, 230601, China
(e-mail: lcl1314@foxmail.com; ahzheng214@foxmail.com; shli0603@foxmail.com).
}

\thanks{
J. Tang and B. Lou are with Anhui Provincial Key Laboratory of Multimodal Cognitive Computation, School of Computer Science and Technology, Anhui University, Hefei, 230601, China 
(e-mail: ahu\_tj@163.com; ahu\_lb@163.com).
}

}

\markboth{Journal of \LaTeX\ Class Files,~Vol.~14, No.~8, August~2021}%
{Shell \MakeLowercase{\textit{et al.}}: A Sample Article Using IEEEtran.cls for IEEE Journals}


\maketitle

\begin{abstract}
Multi-spectral object re-identification (ReID) brings a new perception perspective for smart city and intelligent transportation applications, effectively addressing challenges from complex illumination and adverse weather. However, complex modal differences between heterogeneous spectra pose challenges to efficiently utilizing complementary and discrepancy of spectra information. 
Most existing methods fuse spectral data through intricate modal interaction modules, lacking fine-grained semantic understanding of spectral information (\textit{e.g.}, text descriptions, part masks, and object keypoints).
To solve this challenge, we propose a novel Identity-Conditional text Prompt Learning framework (ICPL), which exploits the powerful cross-modal alignment capability of CLIP, to unify different spectral visual features from text semantics. Specifically, we first propose the online prompt learning using learnable text prompt as the identity-level semantic center to bridge the identity semantics of different spectra in online manner. Then, in lack of concrete text descriptions, we propose the multi-spectral identity-condition module to use identity prototype as spectral identity condition to constraint prompt learning. Meanwhile, we construct the alignment loop mutually optimizing the learnable text prompt and spectral visual encoder to avoid online prompt learning disrupting the pre-trained text-image alignment distribution. In addition, to adapt to small-scale multi-spectral data and mitigate style differences between spectra, we propose multi-spectral adapter that employs a low-rank adaption method to learn spectra-specific features. Comprehensive experiments on 5 benchmarks, including RGBNT201, Market-MM, MSVR310, RGBN300, and RGBNT100, demonstrate that the proposed method outperforms the state-of-the-art methods. The source code is publicly available at \href{https://github.com/lsh-ahu/ICPL-ReID}{https://github.com/lsh-ahu/ICPL-ReID}.
\end{abstract}


\section{Introduction}
\IEEEPARstart{R}e-identification (ReID) aims to match images with the same identity from the gallery repository based on specified query conditions \cite{DBLP:journals/tcsv/LengYT20,DBLP:journals/pami/YeSLXSH22,DBLP:conf/iccv/He0WW0021,DBLP:conf/iccv/ZhengSTWWT15}. With the development of computer vision technology, ReID is increasingly applied within intelligent video surveillance systems. However, challenges such as adverse weather conditions, illumination changes, camera viewpoint variations, and background occlusions still impede the practical application of ReID algorithms. 
As a result, academic and industrial communities have sparked a broad research fervor \cite{DBLP:conf/cvpr/0004GLL019,DBLP:conf/iccv/ZhouYCX19,DBLP:conf/iccv/He0WW0021,Yuan2023SearchingPR,Lu2020DualityGatedMC,Lu2020RGBTTV}, among which introducing auxiliary infrared spectra to complement visual data has drawn growing attention \cite{DBLP:conf/aaai/Li0ZZ020,DBLP:conf/aaai/ZhengWCLT21,Zhong2022GrayscaleEC,Lu2023ModalitymissingRT}.

\begin{figure}
\centering
\captionsetup{font=small}
\includegraphics[width=1.0\linewidth]{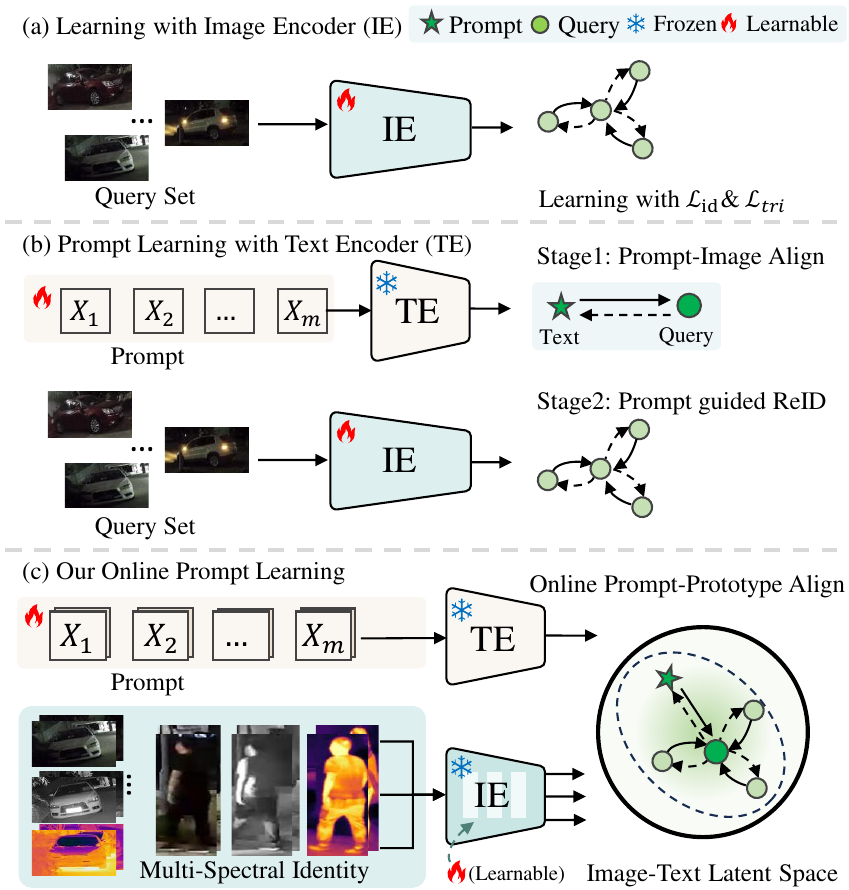}
\caption{(a) Classical pre-training models require ReID learning with id loss and triplet loss \protect\cite{DBLP:conf/cvpr/0004GLL019,DBLP:conf/iccv/He0WW0021}. 
(b) The existing research introduces a two-stage text prompt learning \protect\cite{DBLP:conf/aaai/LiSL23}, which pre-aligned the text prompt for each identity and fine-tuned the ReID task with the text prompt separately. 
(c) Our method proposes an end-to-end text prompt learning framework, seamlessly integrates text prompt learning with multi-spectral ReID task, and alleviates the discrepancies between multi-spectral data. 
}
\label{01_motivation}
\end{figure}

As infrared imaging devices become prevalent in practical production, their unique imaging principle gives people a new perception perspective.
Particularly in low visibility environments such as nighttime, heavy fog, and rainy day \cite{DBLP:conf/mm/WangYCLZ18,DBLP:conf/eccv/SunZYTW18,10098634,DBLP:conf/aaai/ChenCYYDK22}, infrared spectra provides extra discriminative information for identifying challenging queries.
In order to study this problem, Li \textit{et al.} \cite{DBLP:conf/aaai/Li0ZZ020} and Zheng \textit{et al.} \cite{DBLP:conf/aaai/ZhengWCLT21,DBLP:journals/inffus/ZhengZMLTM23} first propose the multi-spectral vehicle and person ReID research tasks and established high-quality benchmarks, extensively promoting research progress in this field. Their contributions underscore the significance of multi-spectral ReID and highlight the challenges encountered in harnessing multi-spectral data effectively.

To effectively utilize multi-spectral data, most existing methods favor a modal-fusion fashion\cite{DBLP:journals/inffus/ZhengZMLTM23,DBLP:conf/aaai/WangHZ024,DBLP:conf/aaai/WangLZLTL24,zhang2024magic}.
However, these methods primarily use visual encoders with intricate model designs to learn modal interactions between spectra, overlooking the fine-grained semantics between infrared and RGB for the same identity, such as text descriptions~\cite{hu2024empowering,DBLP:conf/aaai/ZhaiZH0JC24},
part masks~\cite{DBLP:conf/mm/CuiHSDZW24,DBLP:journals/tifs/HeCWLWJD24,DBLP:conf/cvpr/ChenXJLWW0S23}, and object keypoints~\cite{DBLP:conf/aaai/WangLS0S22,DBLP:conf/mm/LiangJLFWL22}. 
This prior information explicitly guides the model to further focus on the fine-grained semantic features that distinguish identities.
Unlike traditional pre-trained models \cite{DBLP:conf/cvpr/HeZRS16,DBLP:conf/iclr/DosovitskiyB0WZ21,DBLP:conf/cvpr/0003MWFDX22}, CLIP \cite{DBLP:conf/icml/RadfordKHRGASAM21} is trained on large-scale vision-language data, resulting in robust semantic alignment capability. The proposed prompt learning \cite{DBLP:journals/ijcv/ZhouYLL22,zhou2022conditional} further demonstrates its robust generalization and semantic alignment capability across diverse downstream tasks. 
Inspired by these methods, Li \textit{et al.} \cite{DBLP:conf/aaai/LiSL23} and Chen \textit{et al.} \cite{DBLP:conf/mm/0005ZTQX23} propose learning text semantics through learnable text prompts combined with the frozen CLIP text branch. These methods enable the ReID task to effectively capture semantic information from images, even without concrete text labels.

Methods such as CLIP-ReID \cite{DBLP:conf/aaai/LiSL23} and CCLNet \cite{DBLP:conf/mm/0005ZTQX23} propose the two-stage text prompt learning method, exploring the application of image-prompt alignment paradigm in ReID tasks. 
However, as shown in Fig. \ref{01_motivation} (b), the two-stage method requires additional overhead for text prompt pre-alignment, and the first-stage visual branch is frozen to extract multi-spectral features. 
This prevents text prompt from learning spectral-specific semantic features. 
Consequently, the text prompt can only provide fixed semantic constraints during the second-stage of spectra modalities training, preventing the model from aligning to spectra semantics and resulting in suboptimal solutions. 
To address this challenge, as illustrated in Fig. \ref{01_motivation} (c), we propose an end-to-end joint optimization method for text prompt and multi-spectral learning.
By mutually aligning learnable text prompt and optimizing spectral visual encoder in an online text prompt learning manner, 
this method mitigates the semantic shift between text prompt and spectra that occur with separate pre-alignment.

Online text prompt learning is not straightforward.
Due to the lack of concrete text descriptions, the model cannot observe the real spectral and text alignment distribution, and still relies on the pre-trained alignment space of CLIP to align learnable text prompt to spectral modalities. However, direct online spectral alignment inevitably leads to image domain shift when adapting to infrared spectral data which have significant stylistic discrepancies with RGB image, resulting in image modality features deviating from the pre-trained image-text alignment distribution \cite{DBLP:journals/corr/abs-1710-06924,DBLP:conf/nips/SamadhGHKNK023,DBLP:journals/tmm/SongMYFZMW22,DBLP:conf/ijcai/Li0Y0P023}.
To address this problem, we propose the multi-spectral identity condition module to use the multi-spectral identity prototypes as the condition for prompt learning, replacing the spectra instances with these prototypes in online alignment to mitigate the semantic shift after adapting to spectral data, and providing robust multi-spectral identity constraints for text prompt. Additionally, we employ a momentum update method to dynamically aggregate identity prototypes, enabling the learnable text prompt to collaboratively and progressively learn new spectral semantic features during training.

Although online prompt learning has mitigated the semantic shift issue between text prompt and spectral modalities, existing multi-spectral ReID methods still rely on full fine-tuning to learn spectra-specific features. However, due to the smaller scale and significant stylistic discrepancies in multi-spectral data compared to RGB image data, there is a risk of overfitting and dependency on certain spectral modalities \cite{wang2020makes,peng2022balanced,DBLP:conf/ijcai/Li0Y0P023}.
To address these challenges, we propose the multi-spectral adapter, which uses a low-rank adaption method with the lightweight learnable adapter to adapt to different spectra modalities. This approach freezes the original pre-trained model and fine-tunes the few spectra-specific parameters, allowing us to learn spectra features without disturbing the pre-trained image-text alignment distribution.

In summary, we leverage the strong image-text alignment capability of the vision-language pre-training model, aligning multi-spectral modalities with learnable text prompt. By adopting our online prompt learning method, we effectively utilize the spectra data for object ReID tasks. The main contributions of this paper are summarized as follows:
\begin{itemize}
\item We propose a novel online prompt learning framework for the multi-spectral object ReID. To our best knowledge, this is the first work that fully leverages the image-text alignment capabilities of CLIP to enhance the multi-spectral object ReID task.

\item To construct mutual alignment and optimization between text prompt and spectral visual encoder, we propose a multi-spectral identity condition module, which utilizes dynamically updated identity prototypes as constraint condition and constructs alignment loop for prompt learning in the lack of concrete text descriptions.

\item To adapt to multi-spectral data, we propose the multi-spectral adapter, using a low-rank adaptation approach with the lightweight learnable adapter to learn spectral-specific parameters and maintain the pre-trained image-text alignment distribution.

\item To validate the effectiveness of our method, we conduct extensive experiments on five multi-spectral benchmarks, including person and vehicle datasets. The results demonstrated that our method significantly outperformed the state-of-the-art approaches.
\end{itemize}

\begin{figure*}[t]
\centering
\captionsetup{font=small}
\includegraphics[width=1.0\linewidth]{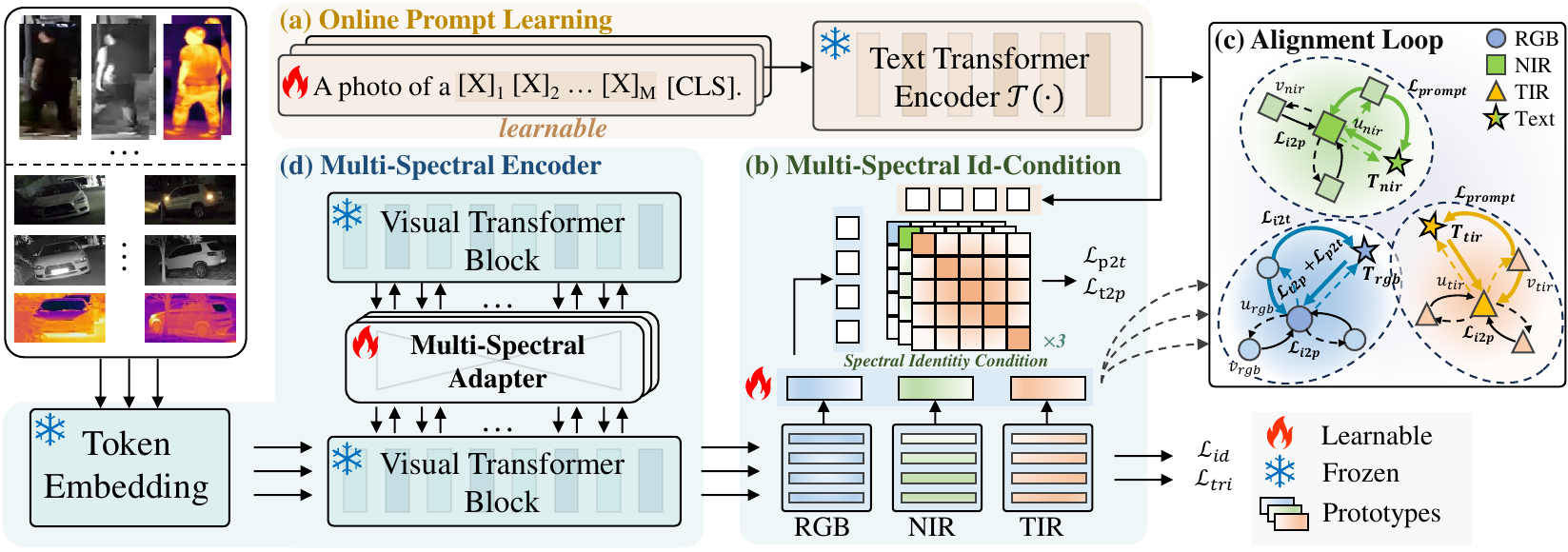}
\caption{Pipeline of our proposed framework. 
(a) For end-to-end training of multi-spectral ReID, online prompt learning leverages learnable text prompt as cross-modal constraints to jointly optimize ReID tasks. 
(b) The multi-spectral identity (id)-condition module first aggregates the RGB, NIR, and  TIR spectral features into identity prototypes, and replaces instance features with prototypes to guide text prompt learning. 
The dynamically updated strategy enables the alignment of spectral-specific semantic features during training.
(c) The alignment loop enables mutual optimization between the text prompts and spectral encoder. It consists of \(\mathcal{L}_{prompt}\) and \(\mathcal{L}_{i2p}\), where \(\mathcal{L}_{t2p} + \mathcal{L}_{p2t}\) within \(\mathcal{L}_{prompt}\) help the text prompt \(T_{m}\) to learn the semantic information of the spectral identity prototype \(u_m\). 
Meanwhile, the image-to-text alignment loss \(\mathcal{L}_{i2t}\), guides the spectral instance \(v_m\) by leveraging semantics without concrete text descriptions.
(d) The multi-spectral encoder freezes and shares most parameters of the visual branch in CLIP for each spectral modality by adding a low-rank adapter to adapt spectra-specific data. 
}
\vspace{-1em}
\label{02_pipeline}
\end{figure*}

\section{Related Work}
\subsection{Multi-Spectral Object ReID}
Multi-spectral object ReID introduces near- and thermal-infrared modalities to enhance the robustness of the model in adverse environments, receiving increasing attention in recent years. 
Unlike single-model object ReID, which requires prior knowledge such as part segmentation, low-light enhancement, and defogging learning to deal with occlusion, nighttime, heavy fog, and domain discrepancy \cite{DBLP:conf/cvpr/LiuZCH019,DBLP:conf/mm/WangYCLZ18,DBLP:conf/eccv/SunZYTW18,10098634,DBLP:conf/aaai/ChenCYYDK22,DBLP:journals/tmm/LiZSL24}, multi-spectral object ReID naturally has the advantage of solving these challenges due to its diversity in imaging principle. 
Although cross-modal object ReID \cite{DBLP:conf/iccv/WuZYGL17,DBLP:conf/mm/ZhengPL00TJ22,DBLP:journals/tmm/LiangJLL23} achieves target retrieval across time and scenes by introducing near-infrared or textual modality \cite{DBLP:conf/aaai/0014LLZZ22,DBLP:conf/mm/LiuZHWZ19,DBLP:conf/ijcai/MiaoLSXY21,DBLP:journals/tmm/FengYCJWLJ23}, it may weaken the unique characteristics of these modalities in the process of eliminating modality discrepancies \cite{DBLP:conf/cvpr/ZhangW23,DBLP:conf/ijcai/LingLLL21}.
To address the above issues, Li \textit{et al.} \cite{DBLP:conf/aaai/Li0ZZ020} pioneer the construction of a multi-spectral vehicle ReID benchmark with visible and infrared modalities as queries, and propose HAMNet to fuse spectra features in a heterogeneity-collaboration aware manner. Zheng \textit{et al.} \cite{DBLP:journals/inffus/ZhengZMLTM23} propose a high-quality multi-spectral vehicle ReID dataset, dubbed MSVR310, covering a broader range of viewpoints, longer time spans, and more environmental complexities, to obtain more consistent multi-spectral feature distributions via cross-directional consistency networks, called CCNet. In the field of person ReID, Zheng \textit{et al.} \cite{DBLP:conf/aaai/ZhengWCLT21} propose the first multi-spectral person ReID dataset, named RGBNT201, and mine complementary features between modalities through progressive fusion with PFNet. Wang \textit{et al.} \cite{DBLP:conf/aaai/WangLZHT22} propose to mine modality-specific features through cross-modal interaction, relationship-based enhanced modality, and multi-modal margin loss.
In contrast to CNN-based methods, the rise of Transformer has brought a new attention mechanism to multi-spectral object ReID. 
Wang \textit{et al.} \cite{DBLP:conf/aaai/WangLZLTL24} propose TOP-ReID, which uses the token permutation module to perform cross-attention fusion of three-spectral features, and propose the complementary reconstruction module to reduce the gap in feature distribution between spectra by reconstructing token-level modal features.
Zhang \textit{et al.} \cite{zhang2024magic} propose an object-centric selection method, called EDITOR, which uses a spatial-frequency token selection module to filter discriminative tokens in each spectra and aggregate token features from different spectra into a multi-spectral representation, providing higher interpretability for multi-spectra ReID task.
Wang \textit{et al.} \cite{DBLP:conf/aaai/WangHZ024} propose HTT, which improves the representation ability of multi-spectral descriptor by constraining the sample distribution spacing between spectra based on ViT, and propose a multi-modal test-time training strategy to improve the model's generalization on unseen test data using self-supervised loss.
However, most multi-modal methods do not consider the identity semantic consistency between spectral modalities. We propose to utilize the vision-language pre-training model CLIP\cite{DBLP:conf/icml/RadfordKHRGASAM21} to mine more discriminative identity semantic features through the prompt learning approach.

\subsection{Vision-Language Pre-training Model}
In recent years, the rise of vision-language pre-training models represented by CLIP \cite{DBLP:conf/icml/RadfordKHRGASAM21,DBLP:conf/icml/0001LXH22}, has increasingly attracted research attention due to their powerful representation and generalization capability on downstream tasks. 
Zhou \textit{et al.} \cite{DBLP:journals/ijcv/ZhouYLL22,zhou2022conditional} propose CoOp and CoCoOp, which achieve excellent transfer performance across various downstream tasks through text learnable prompt. Jia \textit{et al.} \cite{DBLP:conf/eccv/JiaTCCBHL22} and Chen \textit{et al.} \cite{DBLP:conf/nips/ChenGTWSWL22} propose using a few learnable parameters to fine-tune the pre-trained model, achieving fewer resources and more efficient fine-tuning performance. 
However, these methods only explore the common classification tasks, which are distinctly different from ReID task settings. 
CLIP-ReID \cite{DBLP:conf/aaai/LiSL23} proposes using learnable text prompt to guide the learning of ReID task, but the two-stage learning method lacks online alignment between image and text prompt. 
He \textit{et al.} \cite{DBLP:journals/tifs/HeCWLWJD24} combine component segmentation with learnable text prototypes and proposes an adaptive region generation and assessment method to address the challenge of person occlusion ReID, dubbed RGANet. However, this method focuses on the occlusion challenge and relies on additional segmentation supervision signal. 
CCLNet \cite{DBLP:conf/mm/0005ZTQX23} employs learnable text prompt as soft ID labels to contend with the challenges of weak pseudo-label signals and substantial label noise in unsupervised cross-modal person ReID. However, limited by the lack of accurate ID labels, it still cannot dynamically align identity text with image features. 
Different from existing prompt learning methods, we propose using online image-text alignment to transfer pre-trained models to multi-spectral object ReID task more effectively.

\section{Methodology}
In this section, we elaborate on the specific details of the proposed framework, which is trained in an end-to-end fine-tuning manner for multi-spectral object ReID. 
As shown in Fig.~\ref{02_pipeline}, it is comprised of the Online Prompt Learning Strategy (Sec.\uppercase\expandafter{\romannumeral3}.B), Multi-Spectral Identity Condition Module (Sec.\uppercase\expandafter{\romannumeral3}.C), and Multi-Spectral Adapter Module (Sec.\uppercase\expandafter{\romannumeral3}.D).

\subsection{Overview}
First, we introduce the basic pipeline for multi-spectral object ReID training based on CLIP and define relevant symbol definitions. 
Thanks to the image-text contrastive pre-training method and the large-scale training data, the vanilla CLIP model exhibits strong zero-shot capabilities with visual encoder \(\mathcal{I}(\cdot)\) and text encoder \(\mathcal{T}(\cdot)\). 
During training, we freeze them and adopt the parameter-efficient fine-tuning approach to better transfer their generalization ability across diverse visual modalities in multi-spectral object ReID.

In contrast to single- and cross-modal object ReID, multi-spectral object ReID introduces multiple spectra to provide additional auxiliary information. 
Each sample within the query and gallery sets is defined as \(X^i=\{x^i_{rgb}, x^i_{nir}, x^i_{tir}\}\), where \(x^i\) are the \(i\)-th sample containing RGB, Near Infrared (NIR) and Thermal Infrared (TIR) heterogeneous visual modalities. 
Leveraging the Multi-Spectral Identity Condition module, we aggregate image features into identity prototypes \(U^c=\{u^c_{rgb}, u^c_{nir}, u^c_{tir}\}\), where \(u^c\) are multi-spectral cluster centers of the \(c\)-th identity. 
The Online Prompt Learning strategy treats the learnable text modality as identity-level learnable vector \(T^c=\{t^c_{rgb}, t^c_{nir}, t^c_{tir}\}\), where \(t^c\) are text prompt paired with the \(u^c\) for the \(c\)-th identity. 
Conditioned by identity prototypes \(U^c\), we employ text prompt as cross-modality constraints to align image features \(V^i=\{v^i_{rgb}, v^i_{nir}, v^i_{tir}\}\), where \(v^i\) are image features of the \(i\)-th sample.

By fully exploiting the image-text cross-modal alignment capabilities of large-scale pre-trained models, our method avoids designing complex spectral modality interaction modules. During the test inference, only the spectral features need to be concatenated as the final representation.

\begin{figure}[t]
\centering
\captionsetup{font=small}
\includegraphics[width=1.0\linewidth]{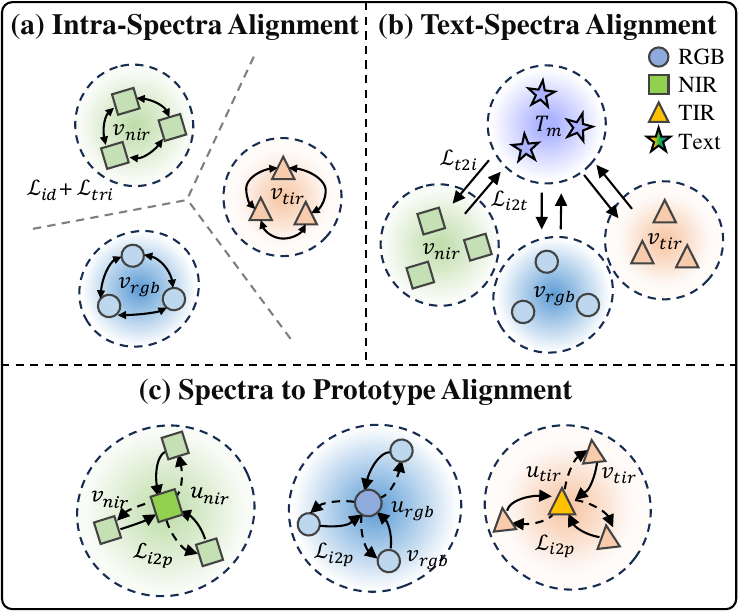}
\caption{
Illustration of traditional instance feature learning strategy. 
(a) The classic ReID metric learning method employs \(\mathcal{L}_{id}\) and \(\mathcal{L}_{tri}\) to enhance intra-class compactness and inter-class separability within the spectra. 
(b) Cross-modal semantic alignment between text and spectra is typically achieved by constructing a latent text-image alignment space with symmetric \(\mathcal{L}_{i2t}\) and \(\mathcal{L}_{t2i}\) losses.
(c) To bring spectral features closer to the prototype and enhance the perception of global sample features within each spectral instance.
}
\label{04_prompt_loss}
\vspace{-1em}
\end{figure}

\subsection{Online Prompt Learning}
Detailed in Fig.~\ref{02_pipeline}, the basic multi-spectral ReID uses a triplet-steam visual encoder \(\mathcal{I}=\{\mathcal{I}_{rgb}, \mathcal{I}_{nir}, \mathcal{I}_{tir}\}\) to extract spectral features, and uses metric learning methods to obtain a compact and separable feature distribution.
However, the heterogeneity between spectra poses challenges for intra-spectra identity alignment. We propose to use learnable identity semantic prompt \(T\) to guide the identity alignment of multi-spectral ReID. The classical prompt learning method CoOp\cite{DBLP:journals/ijcv/ZhouYLL22} defines the text prompt \(T = \) ``\([X]_1, [X]_2,\dots, [X]_M, [CLS]\)'' as a learnable feature, and optimizes prompt with the frozen visual branch. 
However, the training process of ReID is distinctly different from this training strategy.
Typically, we need to unfreeze the visual encoder \(\mathcal{I}\) and learn the fine-grained features of query samples to obtain the final fine-tuned encoder \(\mathcal{I'}\). 
During training, frequently changing image features \(V\) can make it challenging to align the learnable text prompt \(T\). 
One solution is to use a two-stage method to separate the training of text prompt and visual encoder, such as CLIP-ReID\cite{DBLP:conf/aaai/LiSL23}. 
In this method, the frozen visual encoder \(\mathcal{I}\) is used to pre-align the prompt \(T\), and then the frozen prompt \(T\) is used as an unlearnable classifier to optimize the visual encoder \(\mathcal{I'}\). 
However, this approach does not account for the visual encoder \(\mathcal{I}\) changing the original visual feature distribution after adapting to spectral modalities with significant stylistic discrepancies, which broadens the distribution gap between the text prompt \(T\) learned in the first stage and the optimized spectral feature \(V\) in the second stage.

To this end, we propose an online prompt learning training method that collaboratively trains text prompt learning with multi-spectral ReID task. 
Specifically, we define learnable text prompt of each identity described as ``a photo of a \([X]^1_m\), \([X]^2_m\),\dots,\([X]^M_m\), \([CLS]\)", \(m\in[rgb, nir, tir]\), as illustrated in Fig.~\ref{02_pipeline}~(a).
Each learnable token \([X]_m\) is randomly initialized, and \([CLS]\) is object class, (\textit{e.g.}, person or vehicle). Meanwhile, we use the learnable visual encoder \(\mathcal{I}\) to learn the features \(V\) of the ReID task. We use the loss \(\mathcal{L}_{i2t}\) to pull the visual features closer to the text prompt, and the loss \(\mathcal{L}_{t2i}\) to bring text prompt closer to the visual features. The specific formula is as follows:
\begin{equation}
\mathcal{L}_{i2t}=-\frac{1}{M}\log\frac{\exp(\langle v^{c,j}_{m},t^{c}_{m}\rangle/\gamma)}{\sum_{k=1}^{N_m}\exp(\langle v^{c,j}_{m},t^{k}_{m}\rangle/\gamma)},
\label{eq:i2t}
\end{equation}
\begin{equation}
\mathcal{L}_{t2i}=-\frac{1}{M}\log\frac{\exp(\langle t^{c}_{m},v^{c,j}_{m}\rangle/\gamma)}{\sum_{k=1}^{N_m}\exp(\langle t^{c}_{m},v^{k,j}_{m}\rangle/\gamma)},
\label{eq:t2i}
\end{equation}
where \(v^{c,j}_{m}\) is the \(m\)-th spectra feature of the \(j\)-th sample in the \(c\)-th identity, and \(t^{c}_{m}\) is its positive text prompt, \(N_m\) is the number of identities in the \(m\)-th spectra, \(M\) is the number of spectra, and \(\gamma\) is a temperature hyper-parameter.

\subsection{Multi-Spectral Identity Condition}
To avoid online prompt learning disrupting the pre-trained text-image alignment distribution, we propose the Multi-Spectral Identity Condition module. It consists of two components: the Spectral Identity Condition to generate the identity condition by dynamically aggregating instance prototypes, and the Alignment Loop to mutually optimize text prompt and spectral encoder via the identity condition.

\textbf{Spectral Identity Condition.}
Online multi-spectral alignment is challenging due to the lack of concrete text descriptions, making it not a genuine image-text cross-modal task. 
Solely fine-tuning the visual encoder \(\mathcal{I}\) inevitably disrupts the pre-trained image-text alignment distribution, and the significant stylistic discrepancies between different spectra intensify this issue.
To constrain the learning of text prompt, we employ an identity-conditional method during the training process to cluster samples of identical identities into prototypes. 
As shown in Fig.~\ref{02_pipeline}~.(b), it gradually aligns the learnable prompt to the multi-spectral modalities and collaboratively pulls the spectral modalities to the same identity semantic center.
Specifically, before each training epoch, we aggregate image features belonging to the same identity to yield multi-spectral prototypes \(U_{m}=\{u^1_{m}, u^2_{m},...,  u^j_{m}\}\):
\begin{equation}
u_{m}^{c}=\frac{1}{N_{m}^{c}}\sum_{j=1}^{N_{m}^{c}}v_{m}^{c,j}, N_{m}^{c}=|v_{m}^{c}|,
\end{equation}
where \(v_{m}^{c,j}\) are the \(j\)-th instance feature of image sample in the \(c\)-th identity, \(u_{m}^{c}\) are prototypes of the \(c\)-th identity, and \(m\) is a spectral modality in \(\{rgb, nir, tir\}\).

During the training phase, we randomly select \(P\) identities and \(N\) samples for each identity. 
Each sample contains \(M\) different spectra modalities, accumulating to a total of \(P\times N \times M\) images for a mini-batch. 
Subsequently, we apply a momentum update mechanism to dynamically refresh the prototypes of each identity within the memory bank, ensuring that they evolve in sync with the training process.
\begin{equation}
u_{m}^{c,l+1}=\alpha \cdot u_{m}^{c,l}+(1-\alpha) \cdot v_{m}^{c,j},
\end{equation}
where \(l\) and \(l+1\) are the index of the current and next iteration, and \(\alpha\) is the updating factor that controls the feature propagation impacts.

To enhance the robustness of identity prototype, we employ an image-to-prototype alignment loss, denoted as \(\mathcal{L}_{i2p}\), to constrain samples that share the same identity across different spectra:
\begin{equation}
\mathcal{L}_{i2p}=-\frac{1}{M}\log\frac{\exp(\langle v_{m}^{j},u_{m}^{+}\rangle/\gamma)}{\sum_{i=1}^{N_{m}}\exp(\langle v_{m}^{j},u_{m}^{i}\rangle/\gamma)},
\label{eq:i2p}
\end{equation}
where \(u_{m}^{+}\) is the positive feature of \(v_{m}^{j}\), and \(\gamma\) is a temperature hyper-parameter.

\textbf{Alignment Loop.} 
As shown in Fig.~\ref{02_pipeline}~(c), the individual image feature \(V^{i}\) is replaced by identity prototype set \(U^{c}\)\(=\)\(\{u^{c}_{rgb}, u^{c}_{nir}, u^{c}_{tir}\}\).
Owing to the dynamically updated identity prototypes, the learnable text prompt are able to focus on various samples of the same identity. This enables the text prompt to continually observe the latest optimized multi-spectral features during the training process.
Specifically, we employ the contrastive loss \(\mathcal{L}_{t2p}\) and \(\mathcal{L}_{p2t}\) to align the text prompt with the prototypes:
\begin{equation}
\mathcal{L}_{t2p}=-\frac{1}{M}\log\frac{\exp(\langle t^c_{m},u^c_{m}\rangle/\gamma)}{\sum_{k=1}^{N_m}\exp(\langle t^c_{m},u^k_{m}\rangle/\gamma)},
\label{eq:t2p}
\end{equation}
\begin{equation}
\mathcal{L}_{p2t}=-\frac{1}{M}\log\frac{\exp(\langle u^c_{m},t^c_{m}\rangle/\gamma)}{\sum_{k=1}^{N_m}\exp(\langle u^c_{m},t^k_{m}\rangle/\gamma)},
\label{eq:p2t}
\end{equation}
where \(u^c_{m}\), \(t^c_{m}\) are the positive pair of the \(c\)-th identity, and \(\gamma\) is a temperature hyper-parameter.

Intuitively, as depicted in Fig.~\ref{04_prompt_loss}~(a) and Fig.~\ref{04_prompt_loss}~(b), both the object ReID and the text-image alignment tasks require many-to-many feature learning. 
This makes it challenging to effectively learn text prompts and guide the model to recognize identities, particularly in multi-spectral datasets that exhibit low-quality noise and style discrepancies.
However, as shown in Fig.~\ref{04_prompt_loss}~(c), identity prototypes simplify this to a many-to-one problem, effectively alleviating the complexity of prompt optimization.

\begin{equation}
\mathcal{L}_{prompt}=\lambda_1\cdot\mathcal{L}_{i2t}+\lambda_2\cdot(\mathcal{L}_{t2p}+\mathcal{L}_{p2t}),
\label{eq:prompt}
\end{equation}
as shown above Eq.~\eqref{eq:prompt}, we construct the complete alignment loop for online text prompt learning by replacing Eq.~\eqref{eq:t2i} with Eq.~\eqref{eq:t2p} and~\eqref{eq:p2t}. The hyper-parameters \(\lambda_1\) and \(\lambda_2\) are simply set to further smooth the learning of text prompts and their alignment with spectral features during training.
The detailed training pseudo-code is shown in Algorithm \ref{alg:algorithm}.
\begin{figure}[t]
\centering
\captionsetup{font=small}
\includegraphics[width=1.0\linewidth]{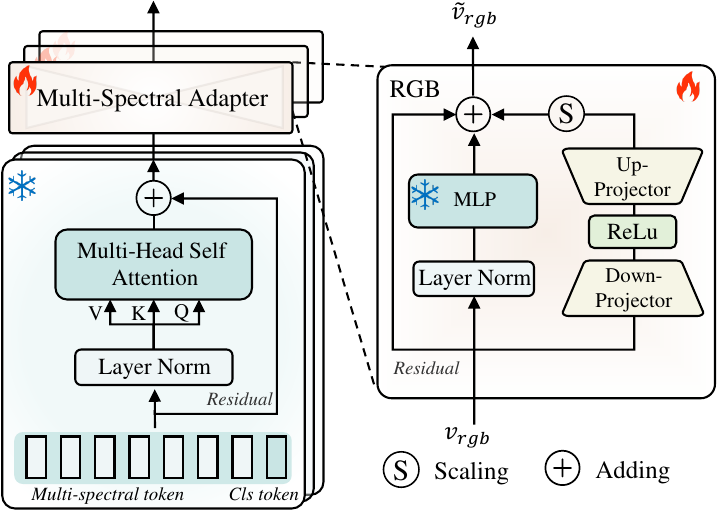}
\caption{Architecture of our multi-spectral adapter.}
\label{03_mma}
\vspace{-1em}
\end{figure}

\subsection{Multi-Spectral Adapter}
Existing pre-trained models primarily focus on RGB images. Although the same object has similar semantics in different spectra, there are still significant discrepancies in stylistic. 
This issue poses challenges for semantic alignment between spectra. 
Additionally, due to the lack of large-scale multi-spectral data, the model may rely on certain spectra when fully fine-tuning on small-scale datasets, and suffer from catastrophic forgetting problems.
To address this challenge, we propose a simple yet effective multi-spectral adaption module that utilizes a low-rank adaption approach to adapt spectra-specific features, as shown in Fig.~\ref{02_pipeline}~(d).

In detail, we first freeze most parameters of the visual encoder \(\mathcal{I}\), preserving only the classification layer, and the last batch normalization layer for training. 
As shown in Fig.~\ref{03_mma}, we introduce a lightweight learnable adapter in the feed-forward network of each transformer block. 
For each adapter, the input features are compressed to the \(\tilde{d}\)-dimensional through the channel down-projection layer, denoted as \(W_{down}\in R^{(d \times \tilde{d})}\). Subsequently, the features are re-expanded to the original \(d\)-dimension through the channel up-projection layer, denoted as \(W_{up}\in R^{(\tilde{d} \times d)}\).
The \(\tilde{d}\)-dimension is the intermediate dimension in the bottleneck layer that is smaller than the \(d\)-dimension. 
A non-linear activation layer ReLU is used to introduce a non-linear transformation for the bottleneck layer between the two linear projection layers. 
Finally, this bottleneck network is connected to the original feed-forward network through residual connections with a scaling factor \(s\). Formal description is as follows:
\begin{equation}
\begin{split}
\tilde{v}_{m}=&s\cdot\mathrm{\textit{ReLU}}(\mathrm{\textit{LN}}(v_{m})\cdot\boldsymbol{W}_\mathrm{down})\cdot\boldsymbol{W}_\mathrm{up}\\
& + \mathrm{\textit{FFN}}(v_{m}) + v_{m},
\end{split}
\end{equation}
where \(\tilde{v}_{m}\) is the optimized spectra feature, which is used as the input for the next block.

\subsection{Optimization}
As in vanilla ReID task setting \cite{DBLP:conf/cvpr/0004GLL019}, we use identity classification loss \(\mathcal{L}_{id}\) to each sample for id constraints, and triplet loss \(\mathcal{L}_{tri}\) to pull together samples sharing the same identity. 
\begin{equation}
\mathcal{L}_{id}= \sum_{i=1}^{N}-q_{i} \log \left(p_i\right) \left\{\begin{array}{ll}
q_i = 0, y \ne i \\
q_i = 1, y=i
\end{array}\right.
\label{eq:idloss}
\end{equation}
\begin{equation}
\mathcal{L}_{tri}=\max(d_p-d_n+\delta,0),
\label{eq:tripleloss}
\end{equation}
where \(y\) as truth ID label and \(p_i\) as ID prediction logits of class \(i\). \(d_p\) and \(d_n\) are feature distances of positive pair and negative pair. \(\delta\) is the margin of triplet loss.

The complete loss function \(\mathcal{L}_{final}\) is defined as follows:
\vspace{-0.2em}
\begin{equation}
\begin{split}
\mathcal{L}_{final}=&\mathcal{L}_{id}+\mathcal{L}_{tri}+\lambda_3\cdot\mathcal{L}_{i2p}+\mathcal{L}_{prompt},
\end{split}
\label{eq:finalloss}
\end{equation}
where \(\lambda_3\) serves as a hyper-parameter to balance the training process.

\begin{algorithm}[tb]
    \caption{Identity-conditional prompt learning process.}
    \label{alg:algorithm}
    \textbf{Input}: Multi-spectral training data \(X_{rgb}\), \(X_{nir}\), \(X_{tir}\). \\
    \textbf{Parameter}: Learnable text tokens \([X]_{rgb}\), \([X]_{nir}\), \([X]_{tir}\), an image encoder \(\mathcal{I}\), a text encoder \(\mathcal{T}\) and update momentum \(\alpha\).\\
    \textbf{Output}: The final multi-spectral loss \(\mathcal{L}_{final}\).
    
    \begin{algorithmic}[1]
        \STATE Initialize \(\mathcal{I}\), \(\mathcal{T}\) from the pre-trained CLIP.
        \FOR{\textit{n} in [1, epochs]}
            \STATE \textcolor{codecomment}{// Extract identity prompt and prototype.}
            \STATE \(T_{rgb}, T_{nir}, T_{tir}=\mathcal{T}([X]_{rgb},[X]_{nir},[X]_{tir})\)
            \STATE \(U_{rgb}, U_{nir}, U_{tir}=average(\mathcal{I}(X_{rgb}, X_{nir}, X_{tir}))\)
            \FOR {\textit{i} in [1, iterations]}
                \STATE \textcolor{codecomment}{// Sample a batch samples from \(X_{rgb}\), \(X_{nir}\), \(X_{tir}\).}
                \STATE \(v_{rgb}, v_{nir}, v_{tir}=\mathcal{I}(x_{rgb}, x_{nir}, x_{tir})\)
                \STATE \textcolor{codecomment}{// Get prompt from \([X]_{rgb}\), \([X]_{nir}\), \([X]_{tir}\).}
                \STATE \(t_{rgb}, t_{nir}, t_{tir}=\mathcal{T}([x]_{rgb}, [x]_{nir}, [x]_{tir})\)
                \STATE Optimize \([X]_1, [X]_2, \dots, [X]_m \) via Eq.~\eqref{eq:t2p} and Eq.~\eqref{eq:p2t}.
                \STATE Optimize visual branch via Eq.~\eqref{eq:i2t} and Eq.~\eqref{eq:i2p}.
                \STATE \textcolor{codecomment}{// Update text prompt and prototype.}
                \STATE \(T^{i+1} = t^{i}\); \(U^{i+1} = \alpha \cdot U^{i} + (1-\alpha) \cdot v^{i}\)
                \STATE Calculate id and triplet loss via Eq.~\eqref{eq:idloss} and Eq.~\eqref{eq:tripleloss}.
            \ENDFOR
        \ENDFOR
    \end{algorithmic}
\end{algorithm}

\section{Experiment}
In this section, we conduct detailed experiments on the proposed framework. First, we introduce the datasets, evaluation protocols, and implementation details (Sec.\uppercase\expandafter{\romannumeral4}.A-B). Second, we conduct comparative experiments with the latest methods on person and vehicle datasets (Sec.\uppercase\expandafter{\romannumeral4}.C). Then, we perform ablation experiments and visual analysis of the proposed method (Sec.\uppercase\expandafter{\romannumeral4}.D). Finally, we further analyze the components of the framework (Sec.\uppercase\expandafter{\romannumeral4}.E).

\subsection{Datasets and Evaluation Protocols}
We conduct experiments on five publicly available multi-spectral datasets, including two multi-spectral person ReID datasets RGBNT201\cite{DBLP:conf/aaai/ZhengWCLT21} and Market-MM\cite{DBLP:conf/aaai/WangLZHT22}, and the multi-spectral vehicle ReID datasets MSVR310\cite{DBLP:journals/inffus/ZhengZMLTM23}, RGBNT100, and RGBN300\cite{DBLP:conf/aaai/Li0ZZ020}.

\textbf{RGBNT201} \cite{DBLP:conf/aaai/ZhengWCLT21} contains 14,361 person images, totaling 4,787 samples, each consisting of 3 spectra modalities, for 201 persons. Within the dataset, 141 identities are divided into the training set, 30 identities into the validation set, and another 30 into the testing set. These samples cover four non-overlapping perspectives. The entire test set is also utilized as a gallery and query set during the testing phase.

\textbf{Market-MM} \cite{DBLP:conf/aaai/WangLZHT22} is a synthetic dataset generated based on the single-modality Market1501~\cite{DBLP:conf/iccv/ZhengSTWWT15}, with a total of 1501 identities and 32,668 sets of multi-spectral samples. 
The training set comprises 751 identities and 12,936 triples, and the rest 750 identities compose the gallery set with 19,732 triples, while the query set contains 750 identities and 3,368 triples. 
To synthesize multi-spectral data, thermal-infrared spectra are generated from RGB images using CycleGAN, near-infrared spectra are created by converting RGB images to grayscale, and RGB images are reduced by 60\% brightness to simulate night scenes.

\begin{table}[ht]
\caption{Comparison performances with the state-of-the-art methods on RGBNT201. 
The best and second-best results are marked in \textbf{bold} and \underline{underline}, respectively.
}
\label{tb:rgbnt201}
\centering
\begin{tabular}{cc|c|llll}
\toprule
& \textbf{Methods} & \textbf{Venue} & \textbf{mAP} & \textbf{R-1} & \textbf{R-5} & \textbf{R-10} \\
\midrule
\multirow{6}{*}{Single} & MUDeep\cite{DBLP:conf/iccv/QianFJXX17} & ICCV17 & 23.8 & 19.7 & 33.1 & 44.3 \\
& MLFN\cite{DBLP:conf/cvpr/ChangHX18} & CVPR18 & 24.7 & 23.7 & 38.5 & 49.5 \\
& PCB\cite{DBLP:conf/eccv/SunZYTW18} & ECCV18 & 32.8 & 28.1 & 37.4 & 46.9 \\
& HACNN\cite{DBLP:conf/cvpr/LiZG18} & CVPR18 & 19.3 & 14.7 & 25.5 & 32.8 \\
& OSNet\cite{DBLP:conf/iccv/ZhouYCX19} & ICCV19 &22.1 & 22.9 & 37.2 & 45.9 \\
& CAL\cite{DBLP:conf/iccv/Rao0L021} & ICCV21 & 27.6 & 24.3 & 36.5 & 45.7 \\
\midrule
\multirow{10}{*}{Multi} & HAMNet\cite{DBLP:conf/aaai/Li0ZZ020} & AAAI20 &27.7 & 26.3 & 41.5 & 51.7 \\
& PFNet\cite{DBLP:conf/aaai/ZhengWCLT21} & AAAI21 & 38.5 & 38.9 & 52.0 & 58.4 \\
& IEEE\cite{DBLP:conf/aaai/WangLZHT22} & AAAI22 & 46.4 & 47.1 & 58.5 & 64.2 \\
& UniCat\cite{DBLP:journals/corr/abs-2310-18812} & NIPSW23 & 57.0 & 55.7 & - & - \\
& HTT\cite{DBLP:conf/aaai/WangHZ024} & AAAI24 & 71.1 & 73.4 & 83.1 & 87.3 \\
& TOP-ReID\cite{DBLP:conf/aaai/WangLZLTL24} & AAAI24 & \underline{72.3} & \underline{76.6} & \textbf{84.7} & \textbf{89.4} \\
& EDITOR\cite{zhang2024magic} & CVPR24 & 66.5 & 68.3 & 81.1 & 88.2 \\
\cmidrule{2-7} 
& CLIP-ReID\cite{DBLP:conf/aaai/LiSL23} & AAAI23 & 71.1 & 71.8 & 80.3 & 85.6 \\
& \textbf{ICPL} & Ours & \textbf{75.1} & \textbf{77.4} & \underline{84.2} & \underline{87.9}
\\	
\bottomrule
\end{tabular}
\end{table}

\begin{table}[ht]
\caption{Comparison performances with the state-of-the-art methods on Market-MM. 
The best and second best results are marked in \textbf{bold} and \underline{underline}, respectively.
Here the superscript \(\ast\) represents the results are reproduced by us.}
\label{tb:marketmm}
\centering
\begin{tabular}{cc|c|llll}
\toprule
& \textbf{Methods} & \textbf{Venue} & \textbf{mAP} & \textbf{R-1} & \textbf{R-5} & \textbf{R-10} \\
\midrule
\multirow{3}{*}{Single} & MLFN\cite{DBLP:conf/cvpr/ChangHX18} & CVPR18 & 42.7 & 68.1 & 87.1 & 92.0 \\
& HACNN\cite{DBLP:conf/cvpr/LiZG18} & CVPR18 & 42.9 & 69.1 & 86.6 & 92.2  \\
& OSNet\cite{DBLP:conf/iccv/ZhouYCX19} & ICCV19 & 39.7 & 69.3 & 86.7 & 91.3 \\
\midrule
\multirow{8}{*}{Multi} & HAMNet\cite{DBLP:conf/aaai/Li0ZZ020} & AAAI20 & 60.0 & 82.8 & 92.5 & 95.0  \\
& PFNet\cite{DBLP:conf/aaai/ZhengWCLT21} & AAAI21 & 60.9 & 83.6 & 92.8 & 95.5  \\
& IEEE\cite{DBLP:conf/aaai/WangLZHT22} & AAAI22 & 64.3 & 83.9 & 93.0 & 95.7 \\
& HTT\cite{DBLP:conf/aaai/WangHZ024} & AAAI24 & 67.2 & 81.5 & 95.8 & 97.8 \\
& \(\text{TOP-ReID}^{\ast}\)\cite{DBLP:conf/aaai/WangLZLTL24} & AAAI24 & 82.0 & 92.4 & 97.6 & 98.6 \\
& \(\text{EDITOR}^{\ast}\)\cite{zhang2024magic} & CVPR24 & 77.4 & 90.8 & 96.8 & 98.3 \\
\cmidrule{2-7} 
& CLIP-ReID\cite{DBLP:conf/aaai/LiSL23} & AAAI23 & \underline{82.5} & \underline{93.7} & \underline{97.9} & \underline{98.8} \\
& \textbf{ICPL} & Ours & \textbf{85.1} & \textbf{94.7} & \textbf{98.4} & \textbf{99.1}
\\
\bottomrule
\end{tabular}
\end{table}

\begin{table}[ht]
\caption{Comparison performances with the state-of-the-art methods on MSVR310. 
The best and second-best results are marked in \textbf{bold} and \underline{underline}, respectively.
}
\label{tb:msvr310}
\centering
\begin{tabular}{c@{}c|c|llll}
\toprule
& \textbf{Methods} & \textbf{Venue} & \textbf{mAP} & \textbf{R-1} & \textbf{R-5} & \textbf{R-10} \\ 
\midrule
\multirow{9}{*}{Single} & DMML\cite{DBLP:conf/iccv/ChenZLZ19} & ICCV19 & 19.1 & 31.1 & 48.7 & 57.2  \\	
& Circle Loss\cite{DBLP:conf/cvpr/SunCZZZWW20} & CVPR20 & 22.7 & 34.2 & 52.1 & 57.2 \\
& PCB\cite{DBLP:conf/eccv/SunZYTW18} & ECCV18 & 23.2 & 42.9 & 58.0 & 64.6  \\
& BoT\cite{DBLP:conf/cvpr/0004GLL019} & CVPRW19 & 23.5 & 38.4 & 56.8 & 64.8 \\
& MGN\cite{DBLP:conf/mm/WangYCLZ18} & MM18 & 26.2 & 44.3 & 59.0 & 66.8  \\
& HRCN\cite{zhao2021heterogeneous} & ICCV21 & 23.4 & 44.2 & 66.0 & 73.9 \\
& OSNet\cite{DBLP:conf/iccv/ZhouYCX19} & ICCV19 & 28.7 & 44.8 & 66.2 & 73.1 \\
& AGW\cite{ye2021deep} & TPAMI21 & 28.9 & 46.9 & 64.3 & 72.3 \\
& TransReID\cite{DBLP:conf/iccv/He0WW0021} & ICCV21 & 26.9 & 43.5 & 62.4 & 70.7 \\
\midrule
\multirow{10}{*}{Multi} & HAMNet\cite{DBLP:conf/aaai/Li0ZZ020} & AAAI20 & 27.1 & 42.3 & 61.6 & 69.5 \\
& PFNet\cite{DBLP:conf/aaai/ZhengWCLT21} & AAAI21 & 23.5 & 37.4 & 57.0 & 67.3 \\
& PFD\cite{DBLP:conf/aaai/WangLS0S22} & AAAI22 & 23.0 &39.9 & 56.3 & 64.0 \\
& FED\cite{DBLP:conf/cvpr/WangZT0HS22} & CVPR22 & 21.7 &37.4 & 58.9 & 67.3 \\
& IEEE\cite{DBLP:conf/aaai/WangLZHT22} & AAAI22 & 21.0 &41.0 & 57.7 & 65.0 \\
& CCNet\cite{DBLP:journals/inffus/ZhengZMLTM23} & INFS23 & 36.4 & 55.2 & 72.4 & 79.7 \\
& TOP-ReID\cite{DBLP:conf/aaai/WangLZLTL24} & AAAI24 & 35.9 & 44.6 & - & - \\
& EDITOR\cite{zhang2024magic} & CVPR24 & 39.0 & 49.3 & - & - \\
\cmidrule{2-7} 
& CLIP-ReID\cite{DBLP:conf/aaai/LiSL23} & AAAI23 & \underline{52.6} & \underline{71.1} & \underline{85.1} & \underline{89.0} \\
& \textbf{ICPL} & Ours & \textbf{56.9} & \textbf{77.7} & \textbf{87.6} & \textbf{91.5}
\\ 
\bottomrule
\end{tabular}
\end{table}

\begin{table}[ht]
\caption{Comparison performances on RGBNT100 and RGBN300. 
The best and second-best results are marked in \textbf{bold} and \underline{underline}, respectively.
Here the superscript \(\ast\) represents the results are reproduced by us.
}
\label{tb:RGBNT100_300}
\centering
\begin{tabular}{c@{}c|c|cc|cc}
\toprule
\multirow{2}{*}{} & \multirow{2}{*}{\textbf{Methods}} & \multirow{2}{*}{\textbf{Venue}} & \multicolumn{2}{c}{\textbf{RGBNT100}} & \multicolumn{2}{c}{\textbf{RGBN300}} \\ 
\cmidrule{4-7} 
& & & mAP & R-1 & mAP & R-1 \\
\midrule
\multirow{5}{*}{Single} & PCB\cite{DBLP:conf/eccv/SunZYTW18} & ECCV18 & 57.2 & 83.5 & 57.7 & 82.0 \\
& MGN\cite{DBLP:conf/mm/WangYCLZ18} & MM18 & 58.1 & 83.1 & 60.5 & 83.7 \\
& ADB\cite{DBLP:conf/iccv/ChenDXYCYRW19} & ICCV19 & 60.4 & 85.1 & 58.9 & 83.1 \\
& OSNet\cite{DBLP:conf/iccv/ZhouYCX19} & ICCV19 & 75.0 & 95.6 & - & - \\
& TransReID\cite{DBLP:conf/iccv/He0WW0021} & ICCV21 & 75.6 & 92.9 & 79.0\(^{\ast}\) & 92.5\(^{\ast}\) \\
\midrule
\multirow{11}{*}{Multi}& HAMNet\cite{DBLP:conf/aaai/Li0ZZ020} & AAAI20 &64.1 & 84.7 & 61.9 & 84.0 \\
& DANet\cite{kamenou2022closing} & ICPR22 & - & - & 71.0 & 89.9 \\
& GAFNet\cite{guo2022generative} & ICSP22 &74.4 & 93.4 & 72.7 & 91.9\\
& GraFT\cite{yin2023graft} & ARXIV23 & 76.6 & 94.3 & 75.1 & 92.1 \\
& GPFNet\cite{he2023graph} & TITS23 & 75.0 & 94.5 & 73.3 & 90.0 \\
& PHT\cite{pan2023progressively} & SENSORS23 & 79.9 & 92.7 & 79.3 & 93.7 \\
& UniCat\cite{DBLP:journals/corr/abs-2310-18812} & NIPSW23 & 81.3 & 97.5 & 80.2 & 92.9 \\
& TOP-ReID\cite{DBLP:conf/aaai/WangLZLTL24} & AAAI24 & 81.2 & 96.4 & 77.7\(^{\ast}\) & 91.9\(^{\ast}\)  \\
& EDITOR\cite{zhang2024magic} & CVPR24 & 82.1 & 96.4 & 75.2\(^{\ast}\) & 90.0\(^{\ast}\) \\
\cmidrule{2-7} 
& CLIP-ReID\cite{DBLP:conf/aaai/LiSL23} & AAAI23 & \underline{87.0}& \underline{96.9} & \underline{85.5}& \underline{94.9} \\

& \textbf{ICPL} & Ours & \textbf{87.0} & \textbf{98.6} &\textbf{87.0}& \textbf{96.3} \\

\bottomrule
\end{tabular}
\end{table}

\textbf{MSVR310} \cite{DBLP:journals/inffus/ZhengZMLTM23} contains 6,261 high-quality vehicle images, divided into 310 different vehicles, with 2,087 samples, each consisting of 3 spectra modalities. The training set includes 155 vehicles and a total of 1,032 samples. The gallery set contains 1,055 samples of the remaining 155 vehicles, while the query set consists of 52 randomly selected vehicles and 591 samples from the gallery set. These samples are captured at long time spans, covering 8 viewpoints around the vehicle and various challenges such as illumination change, shadow, reflection, and color distortion.

\textbf{RGBN300 and RGBNT100} \cite{DBLP:conf/aaai/Li0ZZ020} RGBN300 contains 50,125 sample pairs of 300 different vehicles, each pair containing both RGB and near-infrared modality. Each vehicle is collected by 2 to 8 camera views, with 50 to 200 image pairs. The training set randomly selects 150 vehicles with 25,200 image pairs, the rest 150 vehicles with 24,925 image pairs as the gallery set. From these, 4,985 image pairs are used as the query set. On this basis, RGBNT100 selected 100 vehicles and added 17,250 additional thermal-infrared images to form a three-spectra dataset. This dataset includes 8,675 image triples from 50 vehicles for the training set and 8,575 triples from the other 50 vehicles for the test gallery set. From the test gallery, 1,715 samples are selected to form the query set.

\textbf{Evaluation Protocols.} We use the Cumulative Matching Characteristic (CMC) curve and mean Average Precision (mAP) as evaluation metrics. In MSVR310\cite{DBLP:journals/inffus/ZhengZMLTM23}, as in previous work, we adopt a strict evaluation protocol, which filters out samples with the same identity and time span in the matching results using time labels to avoid easy matching. In RGBNT201\cite{DBLP:conf/aaai/ZhengWCLT21}, Market-MM\cite{DBLP:conf/aaai/WangLZHT22}, RGBNT100 and RGBN300\cite{DBLP:conf/aaai/Li0ZZ020}, we follow the commonly used evaluation protocol as in previous works.

\subsection{Implementation Details}
We resize the images of each spectra to 256x128 (128x256 to maintain the aspect ratio of vehicle) and use random horizontal flipping, padding with 10 pixels, random cropping, and random erasing \cite{DBLP:conf/aaai/Zhong0KL020} as feature enhancement strategies. It is worth noting that data augmentation is not used during the prototype aggregation stage. We select ViT-B/16 as our visual backbone, freezing most parameters in the visual and text branches, while keeping a few learnable parameters, including the learnable text prompt ``\([X]_1, [X]_2,\dots, [X]_M, [CLS]\)'', multi-spectral adapter, classifier, the last batch normalization layer and image-text projection layer in the visual encoder. 
The batch size is set to 64, where 16 identities are randomly selected from each small batch, 4 samples are randomly selected from each identity, and each sample includes 3 images with different modalities. 
We employ the Adam optimizer with a weight decay of 0.0005, momentum of 0.9, and an initial learning rate of 3.5\textit{e}-4. 
The training lasts for 120 epochs, and a warmup strategy is used in the first 10 epochs. 
Linear decay of 0.1 is applied at 30 and 50 epochs, with decay rates of 3.5\textit{e}-5 and 3.5\textit{e}-6. 
All our experiments are conducted on one NVIDIA RTX 4090 using Pytorch.

\begin{table*}[ht]
\caption{
Ablation of different components on RGBNT201, MSVR310, and RGBNT100. We use a triplet-steam CLIP visual encoder as the Baseline. 
Our component splits include the Spectral Identity Condition (SIC) and the Alignment Loop (AL) modules in the Multi-Spectral Identity Condition Module (MS-IC), as well as the Multi-Spectral Adapter Module (MS-A).
}
\label{tb:ablation}
\centering
\resizebox{2.0\columnwidth}{!}{
\begin{tabular}{cccc|llll|llll|llll}
\toprule
& \multicolumn{2}{c}{\textbf{MS-IC}} & \multirow{2}{*}{\textbf{MS-A}} & \multicolumn{4}{c}{\textbf{RGBNT201}} & \multicolumn{4}{c}{\textbf{MSVR310}} & \multicolumn{4}{c}{\textbf{RGBNT100}} \\
& \textbf{SIC} & \textbf{AL} & & mAP & R-1 & R-5 & R-10 & mAP & R-1 & R-5 & R-10 & mAP & R-1 & R-5 & R-10 \\
\midrule
(a) & $\times$ & $\times$ & $\times$ & 71.0 & 71.5 & 81.3 & 86.4 & 49.1 & 65.5 & 82.7 & 85.6 & 85.3 & 96.6 & 97.2 & 97.6 \\
(b) & \checkmark & $\times$ & $\times$ & 72.0 & 73.4 & 80.3 & 85.6 & 52.3 & 71.1 & 84.8 & 89.2 & 86.6 & 96.7 & 97.2 & 97.6 \\
(c) & \checkmark & \checkmark & $\times$ & 73.0 & 75.1 & 82.5 & 86.8 & 55.9 & 76.0 & 86.8 & 90.5 & 86.0 & 97.0 & 97.7 & 98.1 \\
(d) & $\times$ & $\times$ & \checkmark & 72.9 & 73.7 & 81.9 & 87.9 & 55.6 & 77.0 & 87.5 & 91.5 & 85.9 & 98.2 & 98.7 & 98.8 \\
(e) & \checkmark & $\times$ & \checkmark & 72.7 & 75.1 & 82.8 & 87.1 & 56.7 & 77.0 & 86.5 & 90.4 & 85.8 & 98.3 & 98.8 & 98.9 \\
\midrule
(f) & \checkmark & \checkmark & \checkmark & \textbf{75.1} & \textbf{77.4} & \textbf{84.2} & \textbf{87.9} & \textbf{56.9} & \textbf{77.7} & \textbf{87.6} & \textbf{91.5} & \textbf{87.0} & \textbf{98.6} & \textbf{99.0} &	\textbf{99.0} \\
\bottomrule
\end{tabular}
}
\end{table*}

\subsection{Comparison with State-of-the-art Methods}
\textbf{Comparison on RGBNT201 and Market-MM.} Table~\ref{tb:rgbnt201} and Table~\ref{tb:marketmm} report our performance on RGBNT201\cite{DBLP:conf/aaai/ZhengWCLT21} and Market-MM\cite{DBLP:conf/aaai/WangLZHT22} datasets. Clearly, our method has significant advantages over traditional methods and achieves the best performance. 
We extend single-modal CLIP-ReID\cite{DBLP:conf/aaai/LiSL23} to triplet-stream one for fair comparison by replicating the visual encoder three times for three spectra.
Specifically, the triplet-stream CLIP-ReID has achieved comparable performance to existing Transformer-based models, such as TOP-ReID\cite{DBLP:conf/aaai/WangLZLTL24}
, EDITOR\cite{zhang2024magic} 
, HTT\cite{DBLP:conf/aaai/WangHZ024}, etc.
However, the two-stage alignment method does not consider the collaborative alignment of spectral and learnable text prompt, preventing text prompt from learning the new spectral data. 
Therefore, when we adopt the online identity-conditional prompt learning (ICPL), the model can build a mutual alignment loop between learnable text prompt and spectral visual encoder, which enables our model to achieve \textbf{4.0\%/5.6\%} and \textbf{2.6\%/1.0\%} mAP/Rank-1 performance improvement on both datasets compared with CLIP-ReID\cite{DBLP:conf/aaai/LiSL23}.

\textbf{Comparison on MSVR310.}
As shown in Table~\ref{tb:msvr310}, most methods encounter a performance drop when facing the viewpoint variation and long time span challenges in MSVR310\cite{DBLP:journals/inffus/ZhengZMLTM23} dataset. The robust semantic generalization capability of CLIP enables the triplet-stream CLIP-ReID to outperform methods using multi-spectral feature fusion by a large margin.
However, our approach further improves performance, achieving a \textbf{4.3\%/6.6\%} mAP/Rank-1 enhancement over CLIP-ReID\cite{DBLP:conf/aaai/LiSL23}. This demonstrates that our model can effectively focus on viewpoint-invariant identity semantic features through collaborative training with identity semantic prompt.

\textbf{Comparison on RGBNT100 and RGBN300.}
As shown in Table~\ref{tb:RGBNT100_300}, our method achieves the best performance on both datasets. 
Notably, on the RGBNT100\cite{DBLP:conf/aaai/Li0ZZ020} dataset, the mAP of ICPL is comparable to that of CLIP-ReID\cite{DBLP:conf/aaai/LiSL23}, with a \textbf{1.7\%} improvement in Rank-1.
This could be explained by the large number of repeated samples from the same viewpoints in the RGBNT100\cite{DBLP:conf/aaai/Li0ZZ020} dataset, which dilutes the model performance in terms of mAP. 
However, the more challenging Rank-1 metric reflects the better matching ability of ICPL.
On the more complex RGBN300\cite{DBLP:conf/aaai/Li0ZZ020} dataset, while the single-modal TransReID\cite{DBLP:conf/iccv/He0WW0021} model has shown relative effectiveness, both TOP-ReID\cite{DBLP:conf/aaai/WangLZLTL24} and EDITOR\cite{zhang2024magic} fail to achieve the expected performance. In contrast, our method, with end-to-end prompt learning of vehicle textual semantics, further boosts \textbf{1.5\%/1.4\%} in mAP/Rank-1 performance over CLIP-ReID\cite{DBLP:conf/aaai/LiSL23}, highlighting the applicability of our approach to vehicle tasks.

\subsection{Ablation Study}
In this section, we conduct a series of ablation experiments on RGBNT201\cite{DBLP:conf/aaai/ZhengWCLT21}, MSVR310\cite{DBLP:journals/inffus/ZhengZMLTM23}, and RGBNT100\cite{DBLP:conf/aaai/Li0ZZ020} to verify the effectiveness of each component in our proposed framework.
This includes the Spectral Identity Condition (SIC) and the Alignment Loop (AL) within the Multi-Spectral Identity Condition module (MS-IC), and the Multi-Spectral Adapter module (MS-A).

\begin{figure}
\centering
\captionsetup{font=small}
\includegraphics[width=1.0\linewidth]{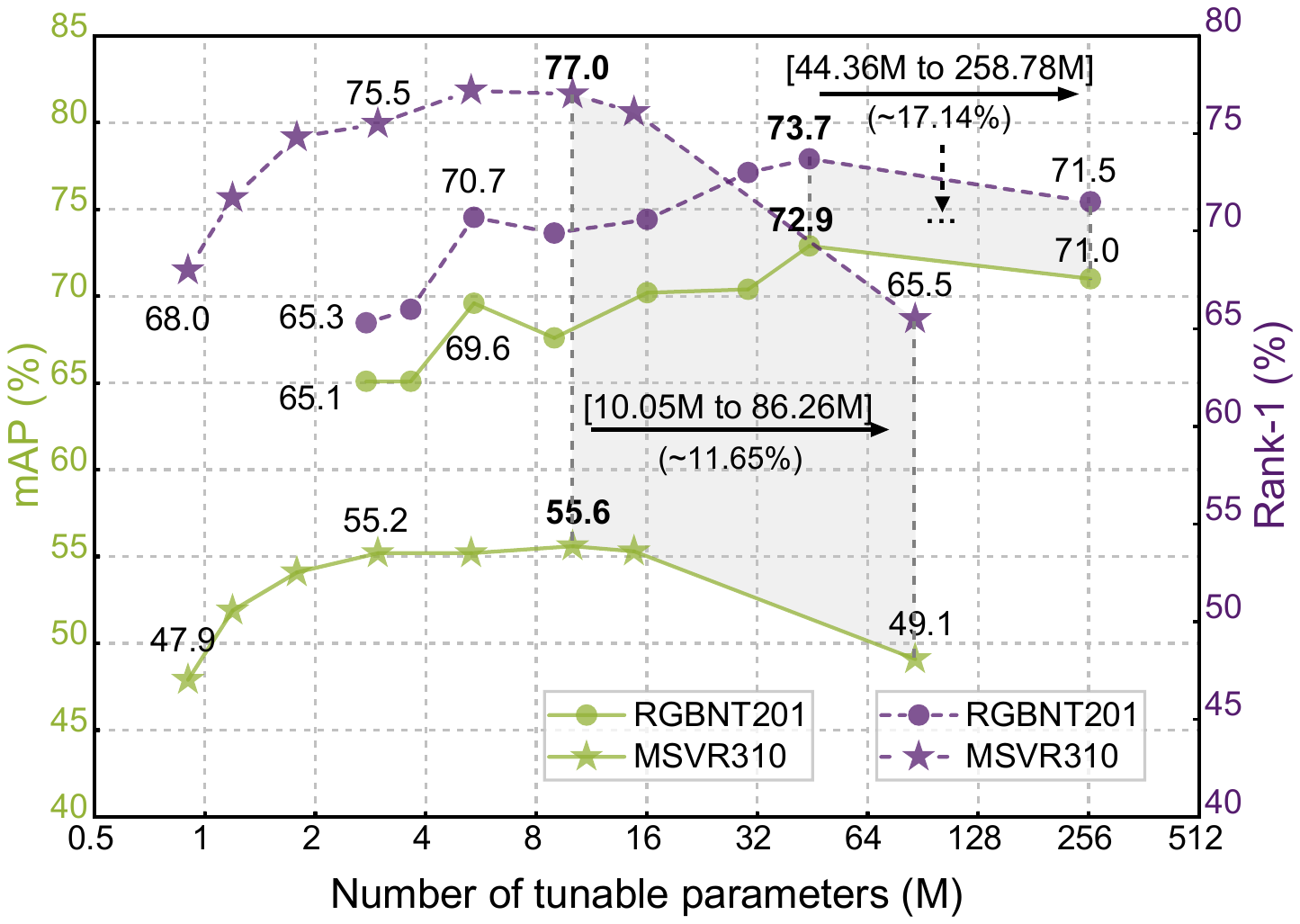}
\caption{The performance trend on mAP and Rank-1 as the number of tunable parameters grows.}
\label{02_adapter}
\end{figure}

\textbf{Ablation Study of Individual Components.}
In order to achieve three spectra inputs, we replicate the CLIP visual encoder three times to form a triplet-stream network, and use it as our baseline in Table~\ref{tb:ablation}(a). We maintain the data augmentation strategy during training and use \(\mathcal{L}_{id}\) and \(\mathcal{L}_{tri}\) as the loss functions. The baseline performance on two datasets highlights the strong representation capability of CLIP. 
When the visual branch is optimized using only the \textbf{SIC} module, the model achieves improvements of \textbf{1.0\%/1.9\%}, \textbf{3.2\%/5.6\%}, and \textbf{1.3\%/0.1\%} in mAP/Rank-1 performance across three datasets, as shown in Table~\ref{tb:ablation}(b).
Prototypes promote spectral feature aggregation using identity anchors, effectively representing sample features within identities. 
As shown in Table~\ref{tb:ablation}(c), introducing a learnable text prompt with the \textbf{AL} module leads to \textbf{2.0\%/3.6\%}, \textbf{6.8\%/10.5\%}, and \textbf{0.7\%/0.4\%} mAP/Rank-1 performance gains across three datasets. 
This improvement is attributed to the robust identity prototype significantly reducing the difficulty of text prompt learning in the absence of real text descriptions, compared with randomly aligned prompts with instance samples.
As shown in Table~\ref{tb:ablation}(d), using the \textbf{MS-A} module results in mAP/Rank-1 performance gains of \textbf{1.9\%/2.2\%}, \textbf{6.5\%/11.5\%}, and \textbf{0.6\%/1.6\%} cross three datasets. 
Notably, the performance improvement on MSVR310 dataset is more significant than on the person dataset. This is due to the MSVR310 dataset encompassing more diverse data in viewpoints and time spans, which benefited our adapter learning with diverse data.
As shown in Table~\ref{tb:ablation}(e), a simple combination of the SIC and MS-A modules leads to a slight performance fluctuation. This phenomenon may be due to the lightweight adapter only relying on the aggregated prototypes, which disrupts the pre-trained feature space distribution, thereby reducing the CLIP generalization performance on unseen test data.
Finally, as shown in Table~\ref{tb:ablation}(f), using all components leads to optimal performance through the mutual optimization of the text prompt and visual encoder, with mAP/Rank-1 performance improvements of \textbf{4.1\%/5.9\%}, \textbf{7.8\%/12.2\%}, and \textbf{1.7\%/2.0\%} cross three datasets, respectively.

\subsection{Further Analysis}
\textbf{Different Variants of Multi-Spectral Identity Condition.} 
To verify the effectiveness of the MS-IC module in online prompt learning, we design different versions of it for comparison. As shown in Table~\ref{tb:ab_i2t}~(a), we replicate the CLIP visual encoder three times as the triplet-stream baseline, which can achieve performance comparable to SOTA methods without the text encoder. 
As shown in Table~\ref{tb:ab_i2t}~(b), We first apply image-to-text loss \(\mathcal{L}_{i2t}\) to the baseline and randomly initialize a text prompt for each identity. 
However, the randomly initialized text prompt lack actual semantics, they cannot effectively assist the model learning, which leads to performance degradation.

\begin{figure}
\centering
\captionsetup{font=small}
\includegraphics[width=1.0\linewidth]{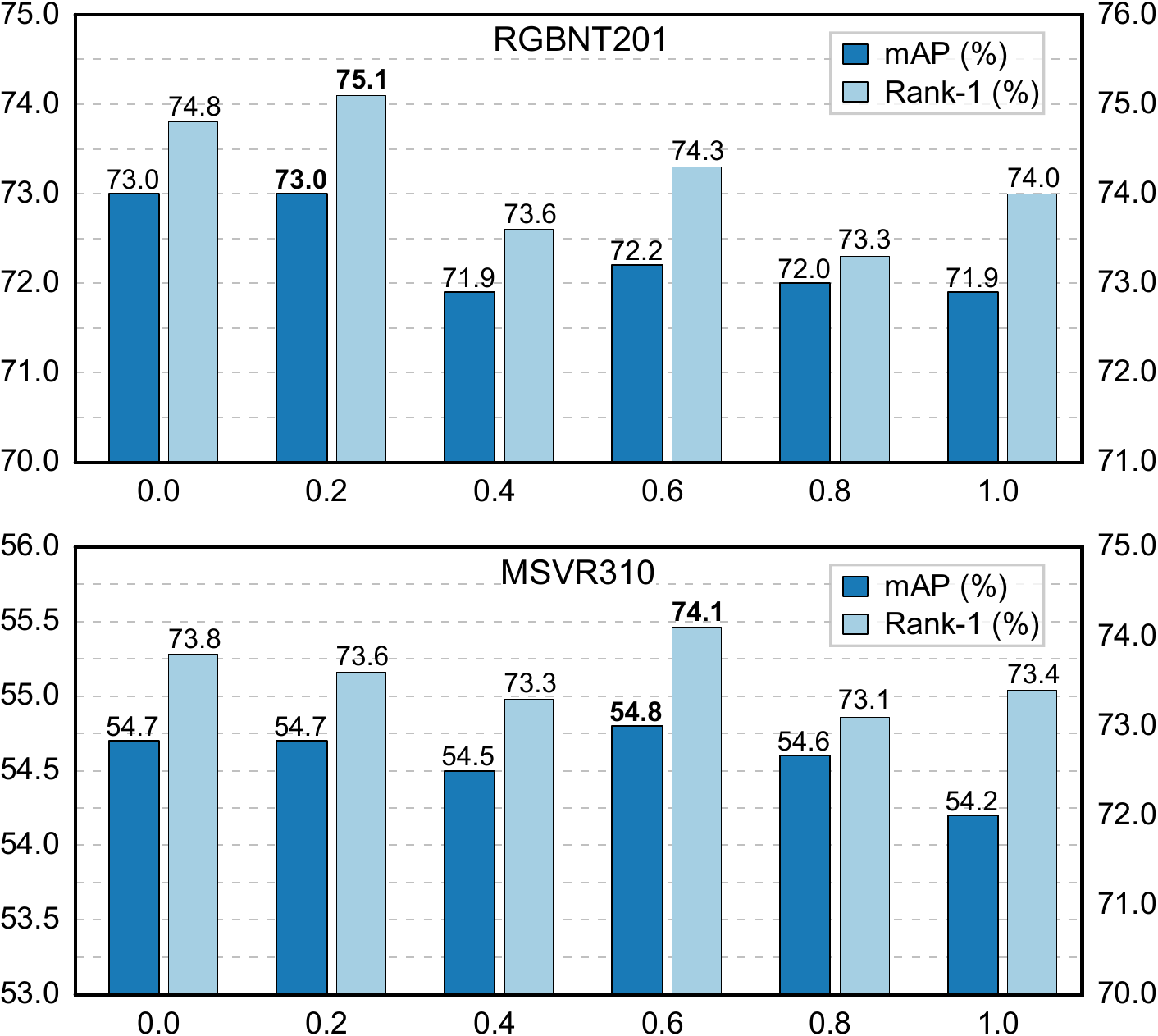}
\caption{Hyper-parameter analysis on prototype factor \(\alpha\).}
\label{01_prototype_alpha_low}
\end{figure}

\begin{table}[ht]
\caption{Different Variants of Multi-Spectral Identity Conditional on RGBNT201 and MSVR310.}
\label{tb:ab_i2t}
\centering
\begin{tabular}{cl|cc|cc}
\toprule
& \multirow{2}{*}{\textbf{Method}} & \multicolumn{2}{c}{\textbf{RGBNT201}} & \multicolumn{2}{c}{\textbf{MSVR310}} \\ 
\cmidrule{3-6} & & mAP & R-1 & mAP & R-1 \\
\midrule
(a) & Baseline & 71.0 & 71.5 & 49.1 & 65.5
\\
(b) & + \(\mathcal{L}_{i2t}\) & 64.3 & 65.7 & 46.2 & 66.2 
\\
(c) & + \(\mathcal{L}_{i2t} + \mathcal{L}_{t2i}\) & 68.0 & 65.6 & 52.2 & 70.4
\\
(d) & + \textbf{MS-IC} (\(\mathcal{L}_{i2p} + \mathcal{L}_{prompt}\))  & \textbf{73.0} & \textbf{75.1} & \textbf{55.9} & \textbf{76.0} 
\\ 
\bottomrule
\end{tabular}
\end{table}

For the second variant in Table~\ref{tb:ab_i2t}~(c), we use image-to-text loss \(\mathcal{L}_{i2t}\) and text-to-image loss \(\mathcal{L}_{t2i}\) to align the randomly initialized text prompt with spectral samples. 
Observing the experimental results, after aligning the text prompt with the spectral modalities, the model has a significant performance improvement on the MSVR310 dataset\cite{DBLP:journals/inffus/ZhengZMLTM23}. 
This indicates that identity-related semantics can be learned through online alignment of text prompt, and this identity semantics is effective in multi-spectral ReID task. 
However, on the RGBNT210 dataset\cite{DBLP:conf/aaai/ZhengWCLT21}, the model is still lower than the baseline, indicating that simple online alignment is relatively suboptimal. 
Finally, by replacing the above alignment loss with image-to-prototype loss \(\mathcal{L}_{i2p}\) and prompt loss \(\mathcal{L}_{prompt}\) in Table~\ref{tb:ab_i2t}~(d), our method achieves superior performance improvements on both datasets.
This proves that online identity-conditional prompt learning can effectively transfer the image-text alignment capability of CLIP to multi-spectral ReID task. 

\begin{figure}
\centering
\captionsetup{font=small}
\includegraphics[width=1.0\linewidth]{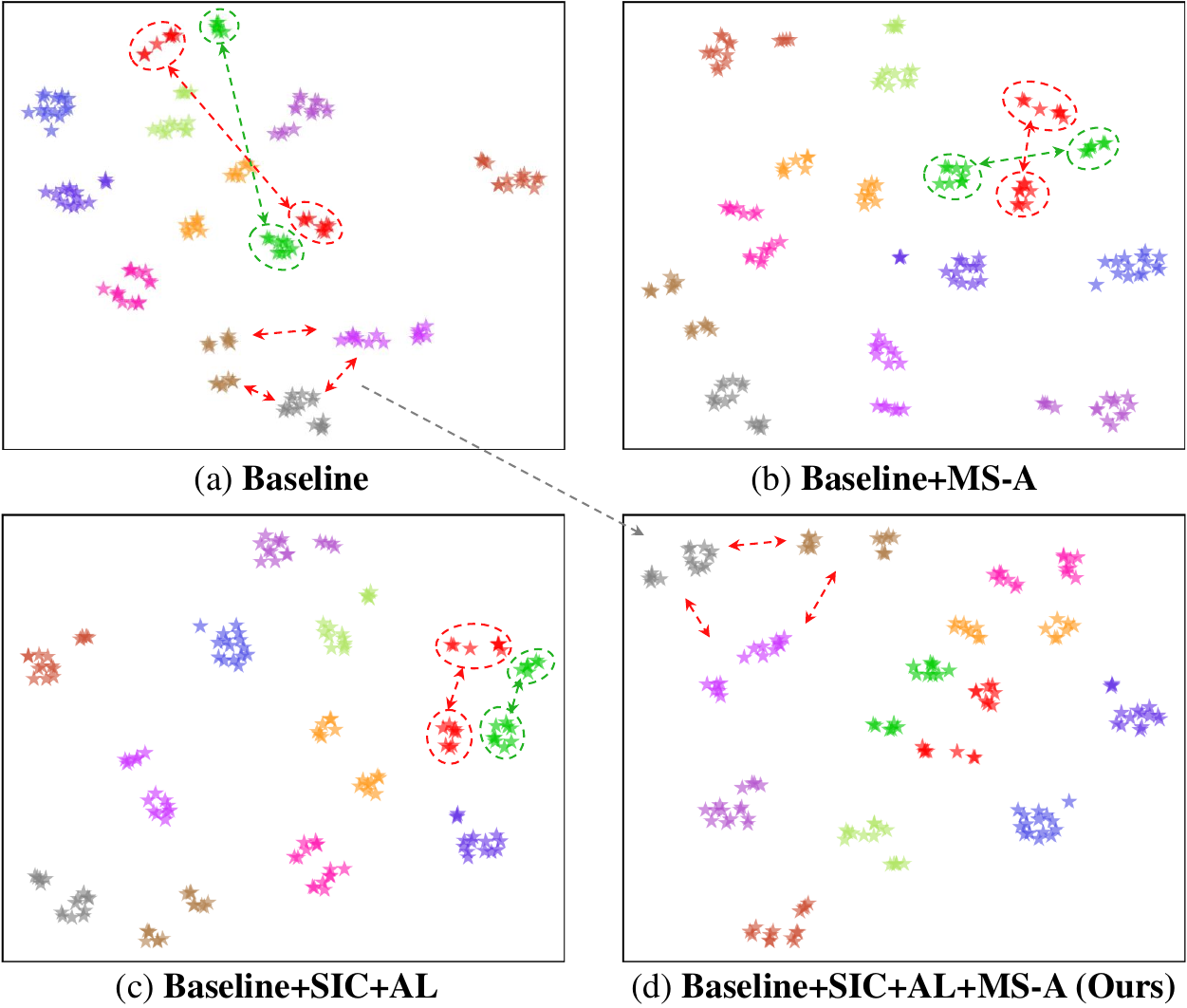}
\caption{Visualization results of the (a) Baseline, (b) Baseline + MS-A, (c) Baseline + SIC + AL, and (d) Baseline + SIC + AL + MS-A (Ours). Better view with colors and zooming in.
}
\label{06_tsne}
\end{figure}

\textbf{Effectiveness on Multi-Spectral Adapter.} 
The low-rank adaption method significantly reduces model training parameters, while effectively addressing discrepancies between the pre-trained model and multi-spectral ReID task. 
By changing the intermediate hidden dimension of adapter, choose from \{16, 32, 64, 128, 256, 512, 768\}, and compare with the full fine-tuning method.
Notably, to further reduce learnable parameters, we validate that different spectra share one learnable adapter on the MSVR310 dataset\cite{DBLP:journals/inffus/ZhengZMLTM23}, as shown in Fig.~\ref{02_adapter}, which also achieves optimal performance with only 10.05M parameters (\(\tilde{d}\)=512), about \(11.65\%\) of the full fine-tuning parameters (86.26M). 
Even when each spectral modality has an individual adapter on RGBNT201, we achieve notable performance with only 44.36M parameters (\(\tilde{d}\)=768), about \(17.14\%\) of the full fine-tuning parameters (258.78M). The above results demonstrate the effectiveness of our multi-spectral adapter in alleviating the discrepancies between multi-spectral data and pre-training data.
Notably, our results indicate that even with the intermediate dimension reduced to 128, the model still performs well on the vehicle dataset. This suggests that the MSVR310 dataset\cite{DBLP:journals/inffus/ZhengZMLTM23} is a curated dataset to provide diverse vehicle samples and rich camera viewpoints while omitting the most redundant vehicle samples, making the dataset sufficiently refined.

\textbf{Hyper-parameters Analysis.} 
During the identity-conditional alignment process, the prototype always plays a pivotal role in maintaining and propagating global sample features to text prompt. As the updating factor increases, the prototypes gradually coagulate from dynamic to static anchor. As shown in Fig.~\ref{01_prototype_alpha_low}, when the factor is less than 1.0, the features of freshly optimized samples are always updated synchronously with the prototypes, enabling the learnable text prompt to adapt to the current training task promptly. In contrast, when the factor is fixed at 1.0, the text prompt cannot dynamically align immediately, resulting in the model falling into sub-optimal solutions. 
This result indicates the significance of learnable text prompt in dynamically aligning to the learned spectra-specific features during the training process.

\begin{table}[ht]
\centering
\caption{Hyper-parameter analysis on Scaling factor \(s\) of adapter.}
\label{tb:scale_factor}
\begin{tabular}{c|cc|cc}
\toprule
\multirow{2}{*}{\textbf{factor}} & \multicolumn{2}{c}{\textbf{RGBNT201}} & \multicolumn{2}{c}{\textbf{MSVR310}} \\
\cmidrule{2-5} & mAP & R-1 & mAP & R-1 \\
\midrule
0.1 & 63.3 & 62.0 & 52.3 & 73.6 \\
0.2 & 66.7 & 67.2 & 55.3 & 76.8 \\
\textbf{0.3} & 68.6 & 68.3 & \textbf{55.6} & \textbf{77.0} \\
0.4 & 70.2 & 69.5 & 55.3 & 77.2 \\
\textbf{0.5} & \textbf{72.9} & \textbf{73.7} & 54.4 & 73.9 \\
0.6 & 70.6 & 70.1 & 55.1 & 75.8 \\
0.7 & 70.5 & 72.2 & 55.0 & 76.3 \\
0.8 & 71.5 & 73.4 & 54.1 & 73.4 \\
0.9 & 70.3 & 72.1 & 54.0 & 74.6 \\
1.0 & 71.2 & 74.2 & 53.1 & 73.3 \\
\bottomrule
\end{tabular}
\end{table}

\begin{table}[ht]
\centering
\caption{
Comparison of Different Learnable Prompt Number M.
}
\label{tb:prompt_num}
\begin{tabular}{c|cc|cc}
\toprule
\multirow{2}{*}{\textbf{num}} & \multicolumn{2}{c}{\textbf{RGBNT201}} & \multicolumn{2}{c}{\textbf{MSVR310}} \\
\cmidrule{2-5} & mAP & R-1 & mAP & R-1 \\
\midrule
1 & 72.7 & 76.6 & 56.2 & 75.0 \\
2 & 74.0 & 77.2 & 55.3 & 75.8 \\
\textbf{4} & \textbf{75.1} & \textbf{77.4} & \textbf{56.9} & \textbf{77.7} \\
8 & 74.1 & 77.5 & 56.6 & 77.0 \\
16 & 73.8 & 74.9 & 55.1 & 75.3 \\
32 & 73.0 & 74.6 & 55.2 & 74.1 \\
\bottomrule
\end{tabular}
\vspace{-1em}
\end{table}

As shown in Table~\ref{tb:scale_factor}, we further explore the influence of the multi-spectral adapter on the frozen visual encoder. 
We fuse the newly learned spectral features of each layer with the original visual features by adding them after adjusting the scale factor. 
When the scale factor is in the middle, the model can achieve a balanced combination of the newly learned spectral features and the original features. However, when the scale factor is too small (\(\leq0.2\)), the model encounters a significant performance decline due to the difficulty in effectively learning new features. On the contrary, when the scaling factor is too large (\(\geq0.9\)), excessive loss of original features leads to performance fluctuation.

\textbf{Comparison of Different Learnable Prompt Number M.}
We analyze the number of learnable prompt tokens, on the RGBNT201\cite{DBLP:conf/aaai/ZhengWCLT21} and MSVR310\cite{DBLP:journals/inffus/ZhengZMLTM23} datasets. As shown in Table~\ref{tb:prompt_num}, an appropriate token number M helps the model achieve optimal performance. 
If M is too small, the limited capacity for semantic learning is insufficient to capture the semantic information from the spectra, leading to a performance decline.
On the other hand, when M is too large, the redundant prompts are hard to optimize, causing the model to experience incorrect semantic guidance that hinders its performance. 

\begin{table*}[ht]
\centering
\caption{Analysis of Loss Weight Factors. 
\(\lambda_1\) for the image-to-text alignment loss \(\mathcal{L}_{i2t}\).
\(\lambda_2\) for the symmetrical prototype-text alignment loss \(\mathcal{L}_{t2p}\)+\(\mathcal{L}_{p2t}\).
and \(\lambda_3\) for the image-to-prototype alignment loss \(\mathcal{L}_{i2p}\). 
The \textbf{bolded} weights represent the default settings.}
\label{tb:loss_anly}
\begin{tabular}{c|cc|c|cc|c|cc||c|cc|c|cc|c|cc}
\toprule
\multicolumn{9}{c||}{\textbf{RGBNT201}} & \multicolumn{9}{c}{\textbf{MSVR310}} \\
\midrule
\(\lambda_1\) & mAP & R-1 & \(\lambda_2\) & mAP & R-1 & \(\lambda_3\) & mAP & R-1 
&
\(\lambda_1\) & mAP & R-1 & \(\lambda_2\) & mAP & R-1 & \(\lambda_3\) & mAP & R-1\\
\midrule
0.0 & 73.8 & 76.7 & 0.0 & 72.8 & 75.6 & 0.0 & 71.2 & 73.1 & 0.0 & 55.0 & 73.6 & 0.0 & 56.2 & 76.1 & 0.0 & 55.0 & 75.0 \\

\textbf{0.1} & \textbf{75.1} & \textbf{77.4} & 0.1 & 73.7 & 76.6 & 0.1 & 72.3 & 74.5 & \textbf{0.1} & \textbf{56.9} & \textbf{77.7} & 0.1 & 56.3 & 75.5 & 0.1 & 55.5 & 75.0 \\
0.2 & 73.8 & 75.2 & 0.2 & 73.3 & 76.3 & 0.2 & 73.0 & 74.8 & 0.2 & 55.9 & 74.3 & 0.2 & 57.0 & 76.5 & 0.2 & 55.4 & 75.0 \\
0.3 & 73.7 & 76.6 & 0.3 & 74.2 & 77.4 & 0.3 & 73.0 & 75.2 & 0.3 & 55.3 & 74.6 & 0.3 & 56.6 & 76.6 & 0.3 & 56.4 & 76.0 \\
0.4 & 73.0 & 75.2 & 0.4 & 74.1 & 77.0 & 0.4 & 73.0 & 74.9 & 0.4 & 55.6 & 75.5 & 0.4 & 56.9 & 76.1 & 0.4 & 56.1 & 76.3 \\
0.5 & 72.9 & 75.0 & 0.5 & 73.5 & 76.6 & 0.5 & 73.5 & 75.8 & 0.5 & 55.7 & 75.5 & 0.5 & 56.7 & 75.6 & 0.5 & 56.7 & 76.0 \\
0.6 & 72.8 & 75.0 & 0.6 & 73.8 & 76.0 & 0.6 & 73.3 & 76.3 & 0.6 & 56.2 & 75.0 & 0.6 & 56.1 & 75.1 & 0.6 & 56.2 & 77.2 \\
0.7 & 72.8 & 74.3 & 0.7 & 73.8 & 75.5 & 0.7 & 73.8 & 76.0 & 0.7 & 55.7 & 75.3 & 0.7 & 56.1 & 76.0 & 0.7 & 56.5 & 75.1 \\
0.8 & 72.6 & 73.9 & 0.8 & 73.5 & 75.8 & 0.8 & 74.4 & 76.4 & 0.8 & 56.6 & 75.6 & 0.8 & 56.4 & 76.1 & 0.8 & 56.9 & 76.6 \\
0.9 & 72.8 & 73.3 & 0.9 & 73.7 & 76.4 & \textbf{0.9} & \textbf{75.1} & \textbf{77.4} & 0.9 & 56.0 & 75.8 & 0.9 & 56.7 & 75.5 & \textbf{0.9} & \textbf{56.9} & \textbf{77.7} \\
1.0 & 73.0 & 74.0 & \textbf{1.0} & \textbf{75.1} & \textbf{77.4} & 1.0 & 74.2 & 76.1 & 1.0 & 56.2 & 76.3 & \textbf{1.0} & \textbf{56.9 }& \textbf{77.7} & 1.0 & 56.8 & 75.8 \\

2.0 & 69.2 & 70.9 & 2.0 & 73.3 & 75.4 & 2.0 & 71.0 & 73.7 & 2.0 & 54.5 & 75.0 & 2.0 & 56.5 & 76.8 & 2.0 & 56.3 & 76.8 \\
5.0 & 61.3 & 62.4 & 5.0 & 72.6 & 75.1 & 5.0 & 69.9 & 71.3 & 5.0 & 52.0 & 72.4 & 5.0 & 56.2 & 76.6 & 5.0 & 54.0 & 74.1 \\
\bottomrule
\end{tabular}
\end{table*}

\begin{table}[ht]
\centering
\caption{
Comparison of the Computation Cost on MSVR310, RGBNT201, and RGBNT100 datasets. All models are evaluated on a single 4090 GPU.
}
\label{tb:efficiency}
\begin{tabular}{c|cc|cc|cc}
\toprule
\multirow{2}{*}{\textbf{Method}} & \multicolumn{6}{c}{\textbf{Training Times \(\downarrow\)}} \\
\cmidrule{2-7} & \multicolumn{2}{c}{20 epoch} & \multicolumn{2}{c}{40 epoch} & \multicolumn{2}{c}{60 epoch} \\ 
\midrule
\multicolumn{7}{c}{(1) MSVR310 (Images: 2678)}\\
\midrule
CLIP-ReID\cite{DBLP:conf/aaai/LiSL23} & 506s & 1x & 861s & 1x & 1212s & 1x \\
ICPL(Ours) & 419s & 0.83x & 820s & 0.95x & 1220s & 1.00x \\
\midrule
\multicolumn{7}{c}{(2) RGBNT201 (Images: 5623)}\\
\midrule
CLIP-ReID\cite{DBLP:conf/aaai/LiSL23} & 754s & 1x & 1273s & 1x & 1769s & 1x \\
ICPL(Ours) & 818s & 1.08x & 1625s & 1.28x & 2439s & 1.38x \\
\midrule
\multicolumn{7}{c}{(3) RGBNT100 (Images: 18965)}\\
\midrule
CLIP-ReID\cite{DBLP:conf/aaai/LiSL23} & 2389s & 1x & 4193s & 1x & 5893s & 1x \\
ICPL(Ours) & 1856s & 0.78x & 3725s & 0.89x & 5585s & 0.95x \\
\bottomrule
\end{tabular}
\vspace{-1em}
\end{table}

\begin{figure*}
\centering
\captionsetup{font=small}
\includegraphics[width=0.95\linewidth]{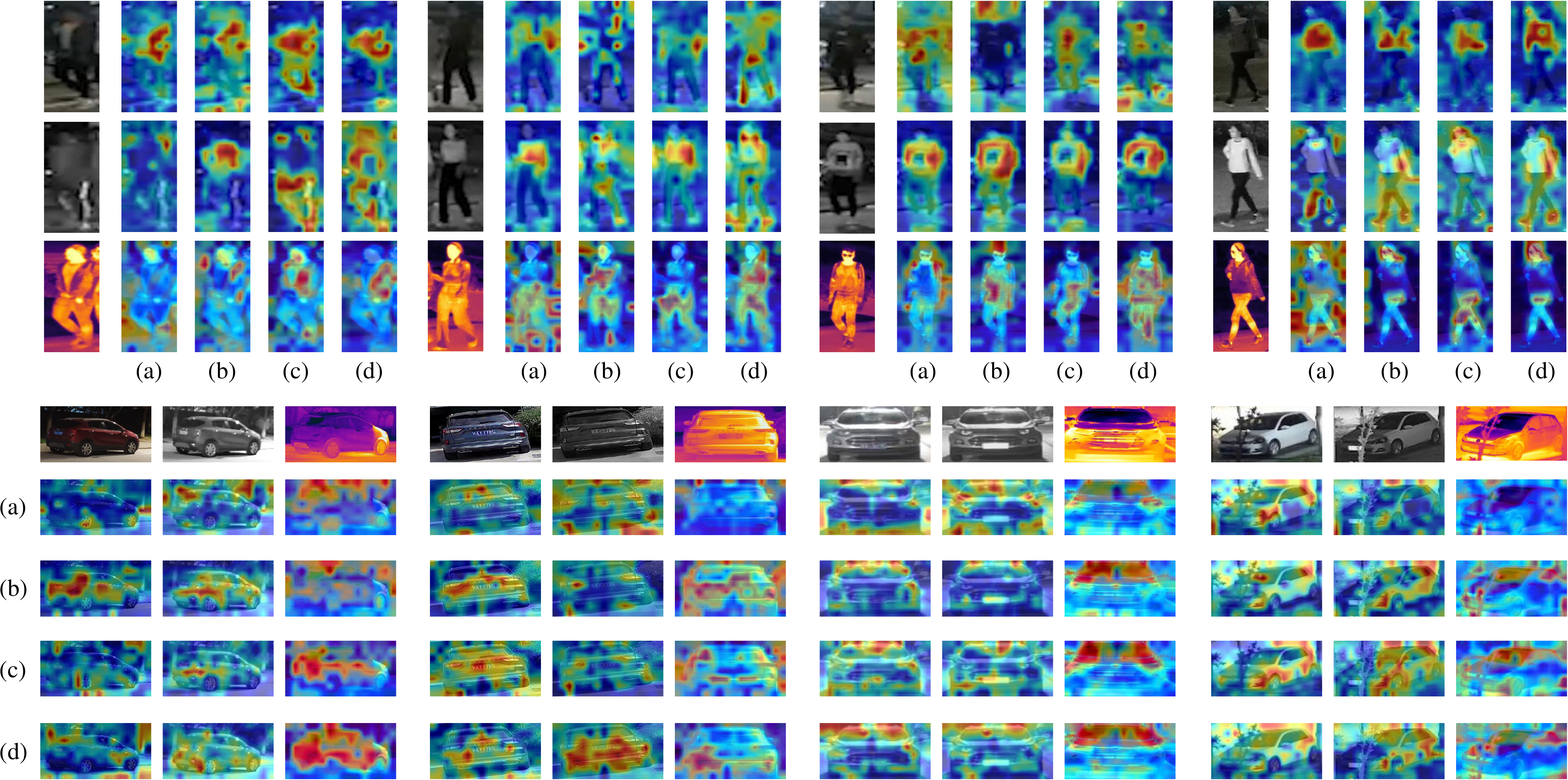}
\caption{Visualization results of the (a) Baseline, (b) Baseline + MS-A, (c) Baseline + SIC + AL, and (d) Baseline + SIC + AL + MS-A (Ours), drawn by Grad-CAM \protect\cite{selvaraju2017grad}. Better view with colors and zooming in.}
\label{04_cam_vis_short}
\vspace{-1em}
\end{figure*}

\begin{figure*}
\centering
\captionsetup{font=small}
\includegraphics[width=0.95\linewidth]{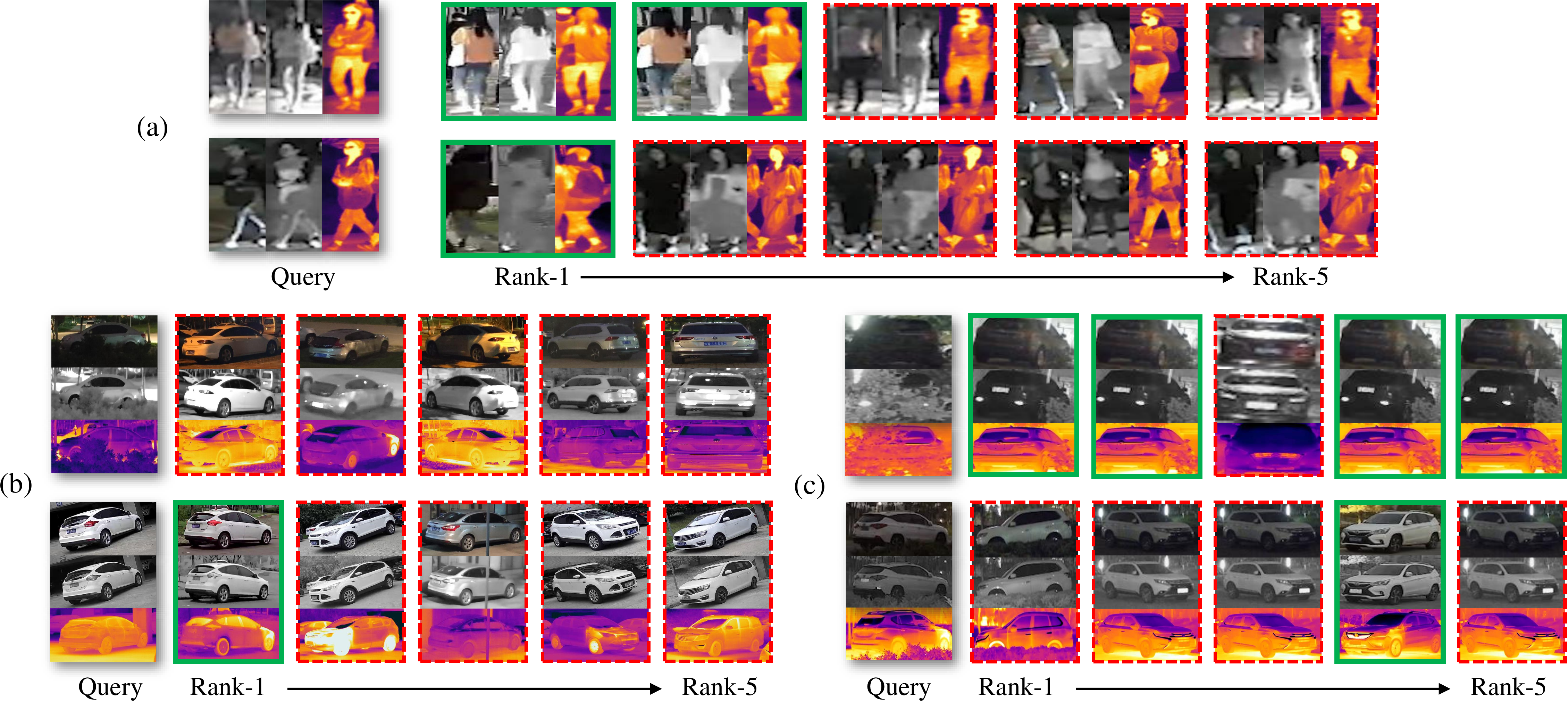}
\caption{
Visualization of top-5 failure cases on (a) RGBNT201\cite{DBLP:conf/aaai/ZhengWCLT21}, (b) MSVR310\cite{DBLP:journals/inffus/ZhengZMLTM23}, and (c) RGBNT100\cite{DBLP:conf/aaai/Li0ZZ020} datasets.}
\label{07_failure_case}
\vspace{-1em}
\end{figure*}

\textbf{Balancing and Trade-offs of Loss Weight Factors.} 
To further insight into the mutual optimization process between text prompt and visual encoder, we analyze the loss function weight factors on the RGBNT201\cite{DBLP:conf/aaai/ZhengWCLT21} and MSVR310\cite{DBLP:journals/inffus/ZhengZMLTM23} datasets.
As shown in Table~\ref{tb:loss_anly}, \(\mathcal{L}_{t2p}+\mathcal{L}_{p2t}\) and \(\mathcal{L}_{i2p}\) perform best with the default weight factor of 1.0 or slightly reduced to 0.9. 
However, the \(\mathcal{L}_{i2t}\) requires a lower weight factor of 0.1, which aligns with our expectations. 
For the randomly initialized semantic prompt, CLIP-ReID employs a pre-alignment stage to ensure the prompt captures identity semantics. 
In contrast, ICPL optimizes both the semantic prompt and spectral encoder together. 
A lower \(\lambda_1\) factor prevents semantic noise from disrupting the spectral encoder during the early alignment process. 
Meanwhile, setting each weight to 0 degrades performance, indicating their necessity in optimization. 
However, when these weights are increased beyond 2.0 to 5.0, we observe a significant performance drop. 
This suggests that excessively large weights skew the optimization process, potentially leading to optimization imbalances and limiting model generalization. 
Based on the above experimental results, setting weights within a moderate range (e.g., around 0.1 to 1.0) effectively balances the optimization objectives within the alignment loop, achieving optimal performance. 
In future work, we will explore more advanced learnable loss tuning methods~\cite{Yuan2023SearchingPR} to more flexibly and cleverly optimize different training objectives in multi-spectral ReID tasks. 

\textbf{Computational Efficiency Analysis.} As shown in Table~\ref{tb:efficiency}, to evaluate the computational efficiency of our proposed framework, we compare ICPL with CLIP-ReID\cite{DBLP:conf/aaai/LiSL23} across three datasets of varying scales. 
In most scenarios, ICPL exhibits faster computational efficiency across different training settings.
While the training time for ICPL is higher on the RGBNT201 dataset, it still achieves acceptable training efficiency due to the faster convergence in shorter periods.
For example, in the 20-epoch setting, the training time is only 1.08 times that of CLIP-ReID\cite{DBLP:conf/aaai/LiSL23}. 
On the larger scale RGBNT100 dataset, ICPL shows a clear advantage.
In the 20-epoch short training setting, the training time is only 0.78 times that of CLIP-ReID\cite{DBLP:conf/aaai/LiSL23}, which is crucial for large-scale datasets. 

\subsection{Visualization}
\textbf{Feature Distribution.} 
To evaluate the impact of different components in ICPL on model performance, we utilize the T-SNE~\cite{van2008visualizing} to visualize the sample distribution, providing in-depth insights into the approach. As shown in Fig.~\ref{06_tsne}~(b), the lightweight MS-A module reduces the distance between challenging samples, enhancing the pre-trained model to better adapt to multi-spectral data. In Fig.~\ref{06_tsne}~(c), the semantic guidance from SIC and AL further compacts the sample distribution within each identity. Finally, as depicted in Fig.~\ref{06_tsne}~(d), the complete ICPL expands the separation between samples from different identities, improving inter-class separability.

\textbf{Discriminative Attention Maps.} To further validate the effectiveness of our proposed components, we use Grad-CAM \cite{selvaraju2017grad} to visualize the features of each spectral modality. 
As shown in Fig.~\ref{04_cam_vis_short} (a), the simply fine-tuned visual encoder cannot effectively focus on the discriminative regions of the object. In scenarios such as nighttime, the model only focuses on the background region and struggles to generate effective feature responses on infrared spectra. 
This indicates that simple fully fine-tuning cannot effectively transfer the pre-trained visual encoder to multi-spectral datasets with significant stylistic discrepancies.
As shown in Fig.~\ref{04_cam_vis_short} (b) to (d), introducing the MS-A module and online text prompt learning with the SIC and AL modules significantly reduces feature response on background areas, while greatly enhancing response on objects in the infrared-spectral modality. 
Based on the above observations, our identity-conditional prompt learning method effectively promotes the model to focus on regions with rich identity semantics for the same identity in different spectral modalities through our mutual optimization online alignment strategy.

\textbf{Failure Cases Analysis.}
To further analyze the retrieval performance of ICPL in real-world scenarios, we visualize the failure cases across three datasets: RGBNT201\cite{DBLP:conf/aaai/ZhengWCLT21}, MSVR310\cite{DBLP:journals/inffus/ZhengZMLTM23}, and RGBNT100\cite{DBLP:conf/aaai/Li0ZZ020}. 
Thanks to the image-text alignment capability of prompt learning, ICPL significantly improves the model performance.
However, as shown in Fig.~\ref{07_failure_case}, extreme lighting degradation, background occlusion, and low-quality noise within the spectral still pose challenges for ICPL in focusing on object semantics in such scenarios. In the future, we plan to explore more robust and fine-grained multi-spectral prompt learning methods to address these challenges. 

\section{Conclusion}
In this paper, we introduce a novel prompt learning framework that harnesses the cross-modal alignment capabilities of the vision-language pre-training model for the multi-spectral ReID task. 
First, our framework enables online prompt learning for multi-spectral ReID, using learnable text prompt as identity-level spectral semantic center to bridge the identity semantics of different spectra.
Second, we propose the multi-spectral identity condition module, which establishes a mutual alignment loop between the text prompt and spectral visual encoder, making the text prompt well-aligned even without concrete spectral text descriptions.
Finally, we propose the multi-spectral adapter module, utilizing a lightweight adapter to optimize the frozen visual encoder, enabling adaptation to new multi-spectral data while preserving the pre-trained image-text alignment distribution of CLIP. 
Extensive experiments on person and vehicle datasets demonstrate the effectiveness of our method. 
In future work, we will explore the fine-grained text prompt to fully exploit the cross-modal alignment capability of the visual-language pre-trained model. 

\bibliographystyle{IEEEtran}
\bibliography{ieee_icpl}{}

\vfill

\end{document}